\documentclass[journal, twoside]{IEEEtran}
\usepackage{enumitem}
\usepackage{graphicx}
\usepackage{amsmath}
\usepackage{textcomp}
\usepackage{makecell}
\usepackage{algorithm}
\usepackage{algorithmic}
\usepackage{color}
\usepackage{amssymb}
\usepackage{multirow}
\usepackage{booktabs}
\usepackage{amsfonts}
\usepackage{xr}

\AddToHook{enddocument/afteraux}{%
    \immediate\write18{cp output.aux main_refs.aux}%
}

\begin{document}

\title{Autonomous Precision Milling of Biological Structures via Generic Anatomical Priors and Active Boundary Perception}

\author{Enduo Zhao, Xiaofeng Lin, Yifan Wang, Yuhan Song, Weihan Li, Saúl Alexis Heredia Pérez, 

and Kanako Harada, \IEEEmembership{Member,~IEEE}
\thanks{This work was supported by Japan Science and Technology Agency (JST) Moonshot Research and Development under Grant JPMJMS2033. \emph{(Enduo Zhao and Xiaofeng Lin are co-first authors.)} \emph{(Corresponding authors: Xiaofeng Lin.)}}
\thanks{This work involved animal subjects in its research. Approval of all ethical and experimental procedures and protocols was granted by the Animal Care and Use Committee, Graduate School of Medicine, University of Tokyo (Approval No. A2023M042-07).}
\thanks{Enduo Zhao is with the School of Biomedical Engineering, Tsinghua University, Beijing, China. (e-mail: endowzhao@mail.tsinghua.edu.cn.)}
\thanks{Xiaofeng Lin, Saúl Alexis Heredia Pérez, and Kanako Harada are with the Center for Disease Biology and Integrative Medicine, Graduate School of Medicine, The University of Tokyo, Tokyo, Japan. (e-mail: lin-xiaofeng@g.ecc.u-tokyo.ac.jp; saulheredia@g.ecc.u-tokyo.ac.jp; kanakoharada@g.ecc.u-tokyo.ac.jp).}
\thanks{Yifan Wang and Weihan Li are with the Department of Mechanical Engineering, Graduate School of Engineering, the University of Tokyo, Tokyo, Japan. (e-mail: wangyifan971125@gmail.com; liweihan1107@g.ecc.u-tokyo.ac.jp).}
\thanks{Yuhan Song is with the Department of Bioengineering, Graduate School of Engineering, the University of Tokyo, Tokyo, Japan. (e-mail: yuhan-song@g.ecc.u-tokyo.ac.jp).}
}

\markboth{IEEE Transactions on Automation Science and Engineering}
{Enduo Zhao \MakeLowercase{\textit{et al.}}: Autonomous Precision Milling of Biological Structures} 

\maketitle

\begin{abstract}
Autonomous precision milling of biological structures is challenged by incomplete knowledge of target geometry, local material thickness, and critical internal boundaries. Subject-specific preoperative models can address geometric and thickness variations, but static models cannot determine boundary status encountered during execution, while repeated target-specific imaging limits scalability. This article presents a state-adaptive autonomous milling framework that assigns complementary roles to generic anatomical priors and active boundary perception. A generic anatomical prior provides conservative global guidance and is transformed through semantic-guided registration and hybrid vision-force calibration into robot-executable guidance for individual targets. As milling approaches uncertain boundaries, the robot actively probes the remaining structure and uses relative stiffness changes to estimate boundary status and structural detachability. A state-adaptive controller governs transitions between active perception and spatially selective incremental refinement, repeating this cycle until the termination criterion is satisfied. Hierarchical experiments on biological surrogates and in vivo mouse cranial window creation demonstrate accurate anatomical prior transfer, reliable boundary adaptation, and autonomous precision milling of biological structures.
\end{abstract}

\begingroup
\renewcommand{\abstractname}{Note to Practitioners}
\begin{abstract}
Machining a hard outer layer without damaging a thin structure beneath it is challenging when individual scans of each workpiece are unavailable. In mouse cranial-window preparation, scanning every specimen is costly and time-consuming, while using a preset milling depth risks damaging the underlying membrane. This creates a trade-off between processing efficiency and safety.

The proposed approach reduces reliance on specimen-specific imaging and fixed milling depths by adapting the machining process to the actual specimen. In experiments, it completed all tested specimens without damage, compared with only one in four using fixed-depth milling. This improvement in reliability comes at the cost of longer processing time due to additional material checks and local adjustments. Practical deployment currently requires firm fixation, as the system cannot compensate for specimen motion such as breathing. Clinical use would require further validation under realistic operating conditions. The approach may also benefit other tasks involving delicate structures beneath a hard surface, including tooth preparation, orthopedic and otologic bone surgery, and automated laboratory sample preparation.
\end{abstract}
\endgroup


\section{Introduction}\label{Sec:introduction}

\begin{figure}[!t]
\centering
\includegraphics[width=0.4\textwidth]{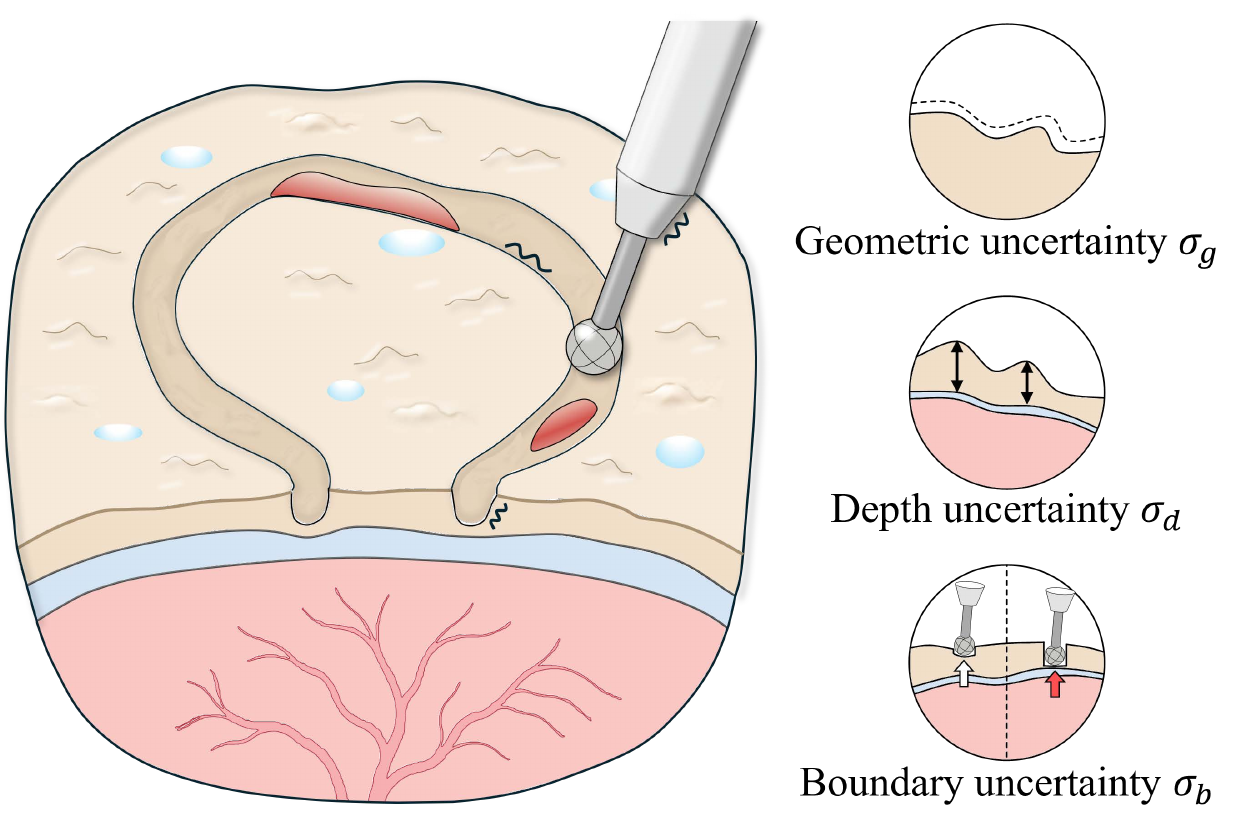}
\caption{Schematic of the three types of uncertainty in autonomous biological milling: geometric uncertainty, depth uncertainty, and boundary uncertainty.}
\label{fig:uncertainty}
\end{figure}

\IEEEPARstart{P}{recision} manipulation is required in a broad range of biomedical procedures, including micro-injection \cite{sugimoto2020microinjection}, micro-grasping \cite{Wang2026TASE}, tissue dissection \cite{strahle2023technical}, and bone micro-processing \cite{dorfer2017novel}, where accurate positioning at small spatial scales, controlled interaction, and minimal tissue damage are essential for reliable experimental outcomes. Furthermore, modern biomedical research increasingly relies on large experimental cohorts \cite{trautmann2025large}, creating a growing need for standardized, scalable workflows that improve reproducibility \cite{von2024systematic} and reduce operator dependence \cite{picozzi2024advances}, thereby motivating robotic automation.

Among these tasks, autonomous precision milling of biological structures (e.g., bone, cartilage, and other hard or mineralized tissues) presents a particularly challenging problem, with applications ranging from cochleostomy \cite{coulson2008autonomous} and precision osteotomy \cite{guo2022total} to hard-tissue sampling \cite{hwang2019robot} and animal craniotomy \cite{marinho2024}. Conventional machining generally operates with a nominally specified workpiece geometry and structure in a controlled environment \cite{russo2024review}; biological milling, in contrast, processes anatomically complex, variable structures in a time-varying biological environment \cite{taylor2003medical}. At the anatomical level, biological targets exhibit substantial spatial and inter-subject variation in surface geometry \cite{maga2017population} and thickness \cite{dai2021vibration}, often with fragile tissue immediately beneath the material being removed. At the environmental level, the operating conditions change during milling. Tissue fluids, blood, and debris can accumulate around the tool, altering the actual milling process \cite{misra2008modeling}. Meanwhile, tissue deformation and physiological motion continuously interfere with stable tool-tissue interaction \cite{takabi2017review}. These variations become particularly consequential near critical interfaces, such as the skull-dura boundary during craniotomy, where a small error can result in unintended penetration and tissue injury \cite{zhan2016craniotomy}.

From the perspective of autonomous execution, these variations result in incomplete knowledge of the actual target structure during milling, giving rise to three task-relevant uncertainties, as illustrated in Fig.~\ref{fig:uncertainty}: geometric uncertainty in the target surface, depth uncertainty in the local material thickness, and boundary uncertainty in the physical state and location of critical internal structures. Together, these uncertainties define the structural information that must be progressively resolved for reliable autonomous milling.

Existing robotic milling approaches commonly reduce geometric and depth uncertainty through subject-specific preoperative anatomical modeling, such as computed tomography (CT) \cite{gui2015novel}, cone-beam computed tomography (CBCT) \cite{fortin2002precision}, micro-computed tomography (micro-CT) \cite{ghanbari2019} and optical coherence tomography (OCT) \cite{navabi25} depending on the application. Such individualized modeling is particularly valuable in safety-critical clinical procedures; however, it is impractical or unnecessary in many biological milling tasks. For example, in large-scale biological research, subject-specific imaging requires additional per-subject preparation and imaging resources \cite{hildebrandt2008anesthesia}, which can limit the scalability of individualized modeling to large experimental cohorts. More importantly, even an accurate preoperative model remains a static description of the target, providing geometric and thickness information but unable to determine the actual boundary status encountered as milling proceeds \cite{han2024review}. Therefore, the fundamental challenge for such tasks is not to construct an increasingly accurate model, but to perceive and respond to the actual structural status during execution.

These observations motivate a division of roles between prior knowledge and online perception for such biological milling tasks, resting on two premises. A generic anatomical prior, a population-averaged 3D model with surface geometry, thickness, and anatomical features, offers conservative guidance on where and approximately how far to mill, whereas online perception determines the precise boundary and whether further milling is required.

Based on this principle, this article proposes a state-adaptive framework for autonomous biological milling. Anatomy-aware registration and hybrid vision-force calibration transform the generic prior into robot-executable guidance, reducing the geometric uncertainty, while open-loop milling removes the bulk of the material and leaves a conservative depth margin that bounds the depth uncertainty. The boundary uncertainty is addressed online through active probing, which yields a target-independent indicator of local boundary status and structural detachability, and the controller then alternates between probing and spatially selective refinement until the remaining structure becomes detachable.

Existing studies laid the groundwork for these two premises but remained limited in their validation and mechanisms. \cite{lin2024} introduced a registration approach in which cranial semantic features anchor a population-level anatomical model to an individual surface. However, it was validated only on static point clouds from post-mortem specimens, and its registration accuracy was insufficient for precision milling. \cite{lin2025object} proposed palpation-based detachability assessment, analogous to the palpation a surgeon performs with a microdrill to judge whether a bone flap is detachable, but relied on an absolute-stiffness threshold specific to the probed object. This work retains these two premises but replaces the underlying mechanisms: the segmentation networks that extract cranial features for registration are redesigned to meet the accuracy the milling task requires, and the boundary assessment is reformulated to rely on relative changes in the measured response during probing, thereby achieving target independence. A finite-state controller coordinates the two mechanisms: the registered population prior provides conservative global guidance, while the boundary state inferred from active probing determines when the remaining safety margin can be released. The resulting framework is progressively validated, from biological surrogates to in vivo cranial window creation.

\begin{table*}[t]
\centering
\caption{Comparison of representative robotic biological milling approaches.}
\label{tab:related_work_comparison}
\small
\setlength{\tabcolsep}{5pt}
\renewcommand{\arraystretch}{1.15}

\begin{tabular}{lcccccc}
\toprule
\multirow{2}{*}{} &
\multicolumn{2}{c}{\textbf{Methodology}} &
\multicolumn{3}{c}{\textbf{Uncertainty Awareness}} &
\multirow{2}{*}{\makecell{\textbf{Experimental}\\\textbf{Target}}} \\
\cmidrule(lr){2-3} \cmidrule(lr){4-6}
& \makecell{Preoperative Planning} & \makecell{Online Perception} & Geometric & Depth & Boundary & \\
\midrule

Gui et al. \cite{gui2015novel}
& Subject-specific & \textemdash & $\checkmark$ & $\checkmark$ & $\times$ & Phantom\\

Ghanbari et al. \cite{ghanbari2019}
& Subject-specific & Passive & $\checkmark$ & $\checkmark$ & $\times$ & In vivo mouse\\

Navabi et al. \cite{navabi25}
& Subject-specific & \textemdash & $\checkmark$ & $\checkmark$ & $\times$ & In vivo mouse\\

Li et al. \cite{li2025optical}
& Subject-specific & \textemdash & $\checkmark$ & $\checkmark$ & $\times$ & In vivo mouse\\

Coulson et al. \cite{coulson2008autonomous}
& \textemdash & Passive & $\times$ & $\checkmark$ & $\checkmark$ & Ex vivo porcine\\

Dai et al. \cite{dai2022vibration}
& \textemdash & Passive & $\times$ & $\checkmark$ & $\triangle^{\dagger}$ & Ex vivo porcine\\

Lin et al. \cite{lin2025object}
& \textemdash & Active & $\checkmark$ & $\checkmark$ & $\checkmark$ & Surrogate \\

Zhao et al. \cite{zhao2025autonomous}
& \textemdash & Passive & $\checkmark$ & $\checkmark$ & $\triangle^{\dagger}$ & Ex vivo mouse\\

Bian et al. \cite{bian2024automatic}
& Subject-specific & Passive & $\checkmark$ & $\checkmark$ & $\triangle^{\dagger}$ & Ex vivo dog/goat\\

Jeong et al. \cite{Jeong2013All}
& Subject-specific & Passive & $\checkmark$ & $\checkmark$ & $\times$ & In vivo mouse\\

\textbf{This work}
& \textbf{Generic} & \textbf{Active} & \textbf{$\checkmark$} & \textbf{$\checkmark$} & \textbf{$\checkmark$} & \textbf{In vivo mouse}\\

\bottomrule
\multicolumn{7}{l}{\footnotesize $^{\dagger}$$\triangle$ indicates awareness of limited task-specific parameters (e.g., milling depth or angle) without integrated closed-loop boundary perception.}
\end{tabular}
\end{table*}

Mouse cranial window creation is selected as the representative in vivo validation task because it naturally combines the three uncertainties discussed above: the skull is thin and spatially nonuniform (0.27--0.51~mm), giving rise to geometric and depth uncertainties, while the fragile underlying dura mater and execution-time deformation introduce boundary uncertainty \cite{Takahashi2024}. Importantly, the framework is formulated for the general problem of autonomous biological milling, in which global anatomical information is incomplete and critical boundaries must be resolved during execution, rather than specifically for this task.

The main contributions of this article are summarized as follows.

\begin{enumerate}

\item A prior-based preoperative planning framework is developed, combining semantic-guided registration with hybrid vision-force calibration to transfer population-level anatomical knowledge to individual targets and generate robot-executable milling trajectories.

\item A state-adaptive hybrid control architecture is proposed to bridge the gap between prior estimation and physical reality. Active boundary perception uses relative stiffness for a target-independent assessment of local mechanical support, and the resulting boundary information drives state transitions and spatially selective refinement until structural detachment.

\item The framework is validated from component-level accuracy to autonomous in vivo execution, demonstrating the effectiveness of state-adaptive control for precision milling under incomplete biological knowledge.

\end{enumerate}

\section{Related Work}\label{Sec:related}

\subsection{Subject-Specific Preoperative Model-Based Planning}

Subject-specific anatomical models have been widely used for preoperative planning of robotic biological milling, providing individualized anatomical information to reduce geometric and depth uncertainty. Dillon et al. \cite{dillon2014preliminary} developed a bone-attached robot for computed CT-guided mastoidectomy. Gui et al. \cite{gui2015novel} proposed a CT-navigation-guided robotic system for pre-planned Le Fort I osteotomy. Liu et al. \cite{liu2024development} demonstrated robotic milling of jawbone cavities based on digital anatomical models. Ghanbari et al. \cite{ghanbari2019} applied micro-CT imaging, while Navabi et al. \cite{navabi25} and Li et al. \cite{li2025optical} explored OCT-based approaches for skull measurement and autonomous planning in mouse craniotomy.

However, such target-specific preoperative imaging provides only static information that cannot address boundary uncertainty, motivating online perception for adaptive milling.

\subsection{Autonomous Milling Based on Online Perception}

An alternative paradigm investigates online perception-based autonomous milling, where boundary uncertainty is addressed during execution. These approaches can be categorized into passive observation and active interaction approaches depending on the source of the signals.

Passive observation-based approaches estimate boundary status by analyzing signals generated during the milling process. Coulson et al. \cite{coulson2008autonomous} used real-time force and torque feedback to detect bony breakthrough for cochleostomy. Dai et al. investigated acoustic \cite{dai2018bioinspired} and vibration signals \cite{dai2022vibration} for tissue-state recognition, while Xia et al. \cite{xia2022vibration} used vibration feedback for depth regulation in robotic bone milling. Zhao et al. \cite{zhao2024autonomous} incorporated vision and force feedback to monitor the progress of mouse craniotomy. While these approaches demonstrate the feasibility of online boundary perception, they primarily rely on indirect process signatures, which can be ambiguous when identifying local boundary transitions in a complex biological environment.

Active interaction-based approaches provide more direct information by physically exciting the target structure. Lin et al. \cite{lin2025object} combined active probing and stereo vision-based depth estimation to assess boundary status and structural detachability. However, this strategy was only validated on surrogate models, leaving its feasibility in biological environments unverified.

Furthermore, several studies have combined the aforementioned online perception and anatomical modeling. Zhao et al. \cite{zhao2025autonomous} integrated stereo vision-based surface reconstruction with progress monitoring, and Jeong et al. \cite{Jeong2013All} used nonlinear optical imaging for surface topography and online feedback for mouse cranial window creation. Bian et al. \cite{bian2024automatic} combined magnetic resonance imaging (MRI) and CT-based preoperative modeling with real-time force-based breakthrough detection for cranium milling. These systems demonstrate that online perception can complement preoperative planning; however, the anatomical representation remains subject-specific and the online perception is predominantly passive or limited to task-specific breakthrough/depth adaptation.

A comparison of representative robotic biological milling approaches, highlighting their methodologies, uncertainty awareness, and experimental targets, is summarized in Table~\ref{tab:related_work_comparison}.

\begin{figure*}[!t]
\centering
\includegraphics[width=0.88\textwidth]{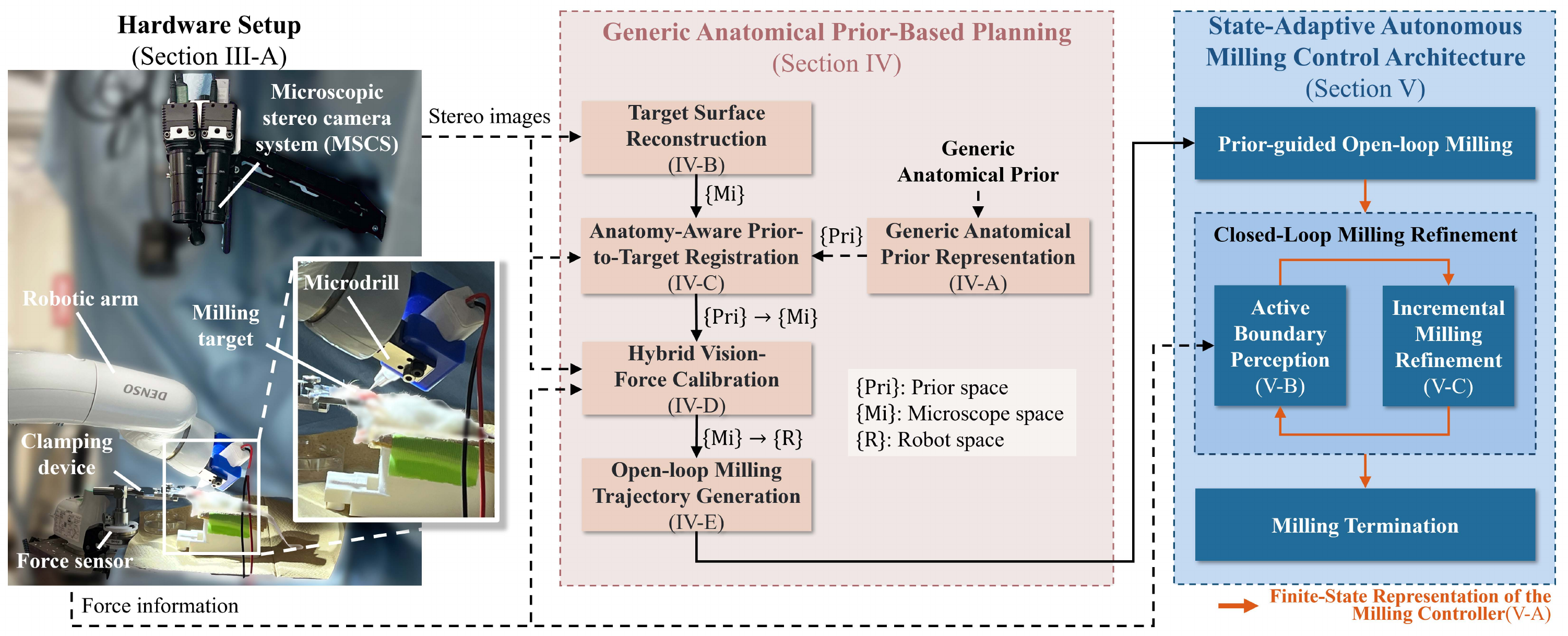}
\caption{Overview of the system. The left side shows the hardware setup, while the right side shows the proposed autonomous robotic precision milling framework. Solid arrows indicate computational and execution flows, whereas dashed arrows represent sensory feedback and prior knowledge transfer.}
\label{fig:overview}
\end{figure*}

\section{System Overview}

\subsection{Hardware Setup}\label{Sec:hard}

The hardware setup of the proposed system is shown in Fig.~\ref{fig:overview}. The system is built upon a robotic platform for scientific exploration \cite{marinho2024}, where a single 8-degree-of-freedom serial manipulator equipped with a microdrill is utilized for precision milling in this study. A microscopic stereo camera system (MSCS) \cite{lin2024} is applied to reconstruct the 3D surface geometry of the target structure within the confined workspace ($20\times30\times10~\text{mm}^3$), providing the geometric information required for prior-to-target and target-to-robot knowledge transformation. In addition, a 6-axis force/torque sensing module is incorporated to acquire tool-target interaction information for active boundary perception.

\subsection{State-Adaptive Autonomous Milling Framework}

The proposed robotic milling framework, as shown in Fig.~\ref{fig:overview}, consists of two interconnected modules: Generic Anatomical Prior-Based Planning (Section~\ref{Sec:preop}) and State-Adaptive Autonomous Milling Control Architecture (Section~\ref{Sec:state-adpative}).

Within the framework, the three task-relevant uncertainties introduced in Section~\ref{Sec:introduction} are given concrete, framework-specific definitions, and each is resolved by a dedicated mechanism, as summarized below.

Geometric uncertainty $\sigma_g$ describes the discrepancy between the prior and actual surface geometries encountered during robot execution. It comprises registration error $\sigma_{g,\text{reg}}$ from prior-to-target registration (Section~\ref{Subsec:alignment}) and calibration error $\sigma_{g,\text{cal}}$ from target-to-robot calibration (Section~\ref{Subsec:Calibration}), both of which the respective modules aim to minimize. The registration and calibration accuracies evaluated in Sections~\ref{subsubsec:registrationexp} and \ref{subsec:calibrationexp} indicate the practical magnitude of these errors and inform the safety margins of the milling trajectory.

Depth uncertainty $\sigma_d$ describes the incomplete knowledge of the local material thickness relative to the generic prior. Because the prior is population-averaged, the true thickness of an individual target deviates from the prior estimate, and $\sigma_d$ characterizes the scale of this deviation. Rather than being estimated online, it is bounded by reserving conservative depth margins during trajectory generation (Section~\ref{Subsec:pathplanner}) and by limiting the per-iteration refinement increment (Section~\ref{Subsec:incremental}).

Boundary uncertainty $\sigma_b$ describes the incomplete knowledge of the physical state and location of critical internal structures. It is characterized by the spatial consistency of the local boundary responses, which active boundary perception (Section~\ref{Subsec:boundaryperception}) measures by probing the remaining structure at each perception cycle. The measured consistency drives the state transitions of the finite-state controller, deciding whether the remaining structure is detachable or requires further refinement (Section~\ref{Subsec:incremental}). $\sigma_b$ is progressively reduced as this perception-refinement cycle iterates.

Overall, the framework follows a hierarchical perception-action strategy: $\sigma_g$ and $\sigma_d$ are resolved during planning, whereas $\sigma_b$ is resolved during execution.

\begin{figure*}[ht]
\centering
\includegraphics[width=0.88\textwidth]{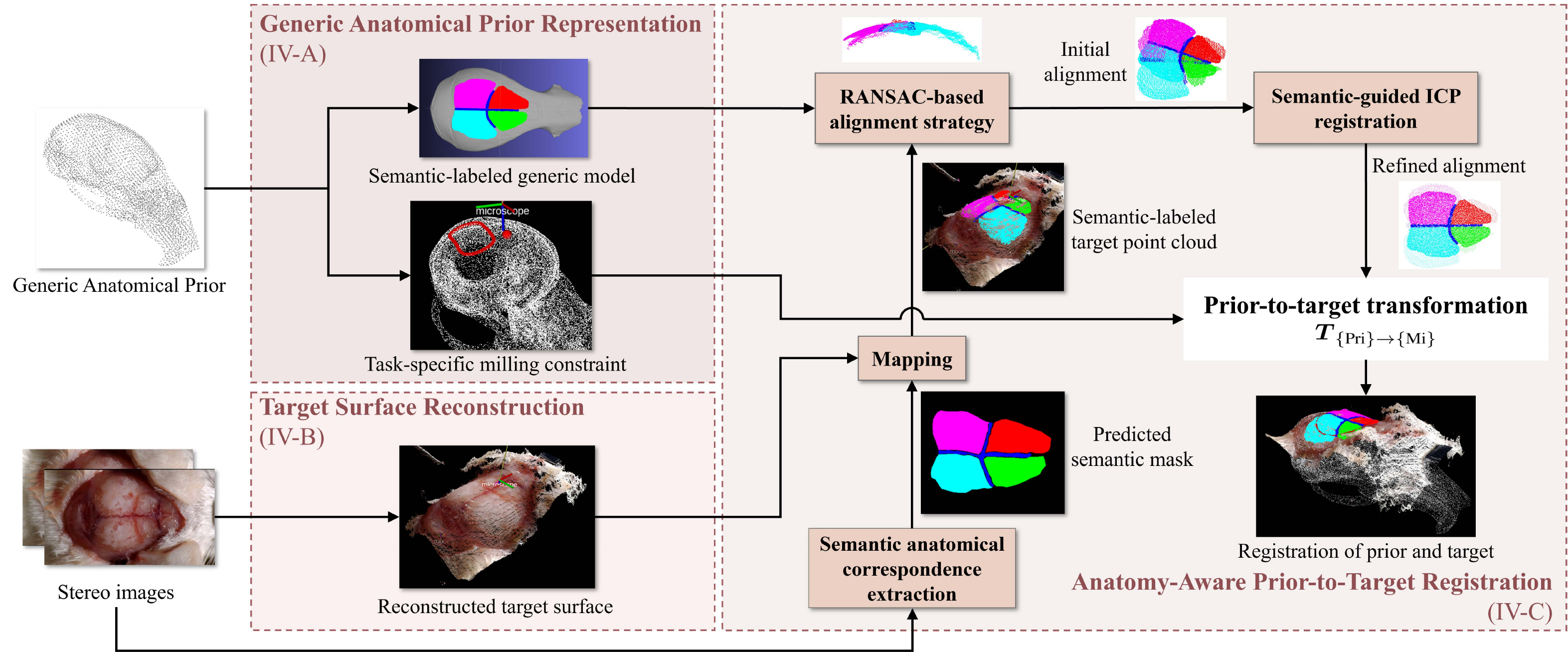}
\caption{
Overview of the proposed workflow from Generic Anatomical Prior Representation, Target Surface Reconstruction to Anatomy-Aware Prior-to-Target Registration.
The prior encodes anatomical geometry and milling constraints, while the target surface is reconstructed using MSCS.
Semantic feature correspondence and registration estimate the transformation between $\{\text{Pri}\}$ and $\{\text{Mi}\}$ for prior-guided milling path transfer.
}
\label{fig:scheme}
\end{figure*}

\section{Generic Anatomical Prior-Based Planning}\label{Sec:preop}

This section presents the Generic Anatomical Prior-Based Planning module, which operates preoperatively to establish prior-to-target and target-to-robot transformations, thereby reducing the initial geometric and depth uncertainty. 

First, the module aligns the generic anatomical prior (Section~\ref{Subsec:preopscanandplanning}) with the reconstructed target surface (Section~\ref{Subsec:reconstruction}) using semantic-guided registration (Section~\ref{Subsec:alignment}). Subsequently, the target-to-robot transformation is estimated via hybrid calibration (Section~\ref{Subsec:Calibration}), enabling the generation of a prior-based target-adapted trajectory for safe, efficient open-loop milling (Section~\ref{Subsec:pathplanner}). Additional implementation details, including mathematical derivations, parameter calibrations, and neural network training protocols, are provided in the Supplementary Materials.

The proposed anatomy-aware registration is built upon \cite{lin2024}, which preliminarily demonstrated the feasibility of registering a generic anatomical prior to point clouds acquired from euthanized mice through semantic correspondences. In this work, we establish a complete mathematical formulation for the registration problem, redesign the semantic correspondence classes to better reflect anatomical significance, and train the model on an expanded dataset. The effectiveness of these improvements is quantitatively evaluated against \cite{lin2024}, as detailed in Section~\ref{subsubsec:registrationexp}.

\subsection{Generic Anatomical Prior Representation}
\label{Subsec:preopscanandplanning}

A generic anatomical prior is introduced to represent transferable anatomical knowledge for robotic milling, which is derived from population-level anatomical characteristics encoded with task-specific constraints.

The generic anatomical prior is represented in the prior coordinate frame $\{\text{Pri}\}$, denoted as

\begin{equation}
\mathcal{P}_{\text{prior}}
=
\left\{\mathcal{A},\mathcal{T}\right\},
\end{equation}

\noindent where $\mathcal{A}$ denotes population-level anatomical information (introduced in Section~\ref{subsubsec:semantic}) and $\mathcal{T}$ represents the task-specific milling constraints (introduced in Section~\ref{subsubsec:startpathfitting}), as illustrated in Fig.~\ref{fig:scheme}. 

For concreteness, we use an open-source population-based mouse skull model provided by \cite{maga2017population} as a representative anatomical realization of the generic prior, which is averaged over 30 female mice at 8 weeks. The cranial window creation task is then considered as the demonstrated task for defining the corresponding task-specific constraints.

\subsubsection{Definition of anatomical information}\label{subsubsec:semantic}

The anatomical component of $\mathcal{P}_{\text{prior}}$ is defined as

\begin{equation}
\mathcal{A}
=
\left\{
\mathbf{S},
\mathbf{D},
\mathbf{L}
\right\},
\label{eq:ASDL}
\end{equation}

\noindent where $\mathbf{S}$ and $\mathbf{D}$ represent the population-level surface geometry and structural thickness distribution.

$\mathbf{L}$ denotes semantic anatomical labels that encode identifiable anatomical regions for establishing correspondence between the prior and individual targets. Specifically, cranial sutures provide positional references due to their anatomical uniqueness, while characteristic curved-surface regions offer complementary orientation constraints. Together, these semantic constraints minimize geometric ambiguity, enabling robust registration despite inter-subject geometric variations.

\subsubsection{Definition of task-specific milling constraints}
\label{subsubsec:startpathfitting}

The task-specific component of $\mathcal{P}_{\text{prior}}$ is defined as

\begin{equation}
\mathcal{T}
=
\left\{
\mathbf{P},
\mathbf{{D}_{P}}
\right\},
\label{eq:TPDP}
\end{equation}

\noindent
where $\mathbf P$ represents the desired milling contour defined as a sequence of $N$ points sampled on the surface geometry $\mathbf{S}$:

\begin{equation}
\mathbf{P}
=
\left\{
\boldsymbol p_{i}^{\{\text{Pri}\}}
\in\mathbb{R}^{3}
|
i=1,\cdots,N
\right\},
\label{eq:Ppri}
\end{equation}

\noindent
where $\boldsymbol p_{i}^{\{\text{Pri}\}}$ denotes the $i$-th point along the contour path, and $N$ determines the spatial resolution of the path. 

$\mathbf{{D}_{P}}$ denotes the local thickness profile obtained by sampling the anatomical thickness distribution $\mathbf{D}$ along the desired path: 

\begin{equation}
\mathbf{{D}_{P}}
=
\begin{bmatrix}
d_1,d_2,\cdots,d_N
\end{bmatrix}^\text{T},
\label{eq:Dp}
\end{equation}

\noindent
where $d_i$ denotes the estimated local structural thickness associated with $\boldsymbol p_{i}^{\{\text{Pri}\}}$.

Together, $\mathbf{P}$ and $\mathbf{{D}_{P}}$ constitute the task-specific milling constraint $\mathcal{T}$ for subsequent trajectory generation.

\subsection{Target Surface Reconstruction}
\label{Subsec:reconstruction}

To obtain individual geometric information, reconstruction of the target surface is performed using the MSCS. The captured stereo images are processed through stereo correspondence and disparity estimation to obtain depth information, which is converted into a 3D point cloud in the microscopic coordinate frame $\{\text{Mi}\}$, as shown in Fig.~\ref{fig:scheme}.

As the reconstructed surface serves as a target-side representation for subsequent registration and normal-direction interaction, local surface normals are estimated from the reconstructed point cloud. Accordingly, the target surface is represented as a geometric map:

\begin{equation}
\mathcal{X}_\text{surf}
=
\left\{
(\boldsymbol q_i^{\{\text{Mi}\}},
\boldsymbol n_i^{\{\text{Mi}\}})
\mid
i=1,\cdots,K
\right\},
\label{eq:smi}
\end{equation}

\noindent
where $\boldsymbol q_i^{\{\text{Mi}\}}\in\mathbb{R}^{3}$ denotes the $i$-th reconstructed surface point, $\boldsymbol n_i^{\{\text{Mi}\}}\in\mathbb{R}^{3}$ represents its corresponding surface normal vector, and $K$ denotes the resolution of the surface.

\subsection{Anatomy-Aware Prior-to-Target Registration}
\label{Subsec:alignment}

Due to the large geometric discrepancy between the surface of the generic anatomical prior $\mathbf{S}$ and the reconstructed surface of the individual target $\mathcal{X}_\text{surf}$, a spatial correspondence between $\{\text{Pri}\}$ and $\{\text{Mi}\}$ is established by exploiting semantically consistent anatomical features across individuals, rather than relying on conventional surface geometry-based registration.

Specifically, the target surface is segmented using the same semantic label scheme $\mathbf{L}$ as the generic prior. As shown in Fig.~\ref{fig:scheme}, a Mask2Former-based multi-class semantic segmentation network \cite{cheng2022masked} is employed to identify predefined anatomical regions for the MSCS observation, and the predicted semantic masks are projected onto the reconstructed surface to generate a semantic-labeled target point cloud. 

The registration is initialized using Random Sample Consensus (RANSAC) \cite{fischler1981random} and subsequently refined by semantic-guided Iterative Closest Point (ICP) \cite{Park2017}, where correspondence searching is restricted to identical anatomical classes. The prior-to-target transformation is obtained by minimizing the alignment error between semantically consistent anatomical correspondences:

\begin{equation}
\boldsymbol{T}_{\{\text{Pri}\}\rightarrow\{\text{Mi}\}}
=
\arg\min_{\boldsymbol{T}}
\sum_{i=1}^{N_r}
\left\|
\tilde{\boldsymbol q}_{\text{sem},i}^{\{\text{Mi}\}}
-
\boldsymbol T
\tilde{\boldsymbol p}_{\text{sem},i}^{\{\text{Pri}\}}
\right\|^{2},
\label{eq:TpritoMi}
\end{equation}

\noindent
where $\tilde{\boldsymbol p}=[\boldsymbol p^\text{T},1]^\text{T}$ denotes the homogeneous representation of the 3D position vector, $N_r$ denotes the number of matched anatomical correspondences, and $\boldsymbol p_{\text{sem},i}^{\{\text{Pri}\}}$ and $\boldsymbol q_{\text{sem},i}^{\{\text{Mi}\}}$ denote matched anatomical points obtained through semantic-guided correspondence estimation.

This registration compensates for the geometric discrepancy between the generic prior and the individual target, minimizing the registration error $\sigma_{g,\text{reg}}$ of the geometric uncertainty $\sigma_g$. The registration error is quantified by the Root Mean Square Error (RMSE) of the minimized residual:

\begin{equation}
\sigma_{g,\text{reg}}
=
\sqrt{
\frac{1}{N_r}
\sum_{i=1}^{N_r}
\left\|
\tilde{\boldsymbol q}_{\text{sem},i}^{\{\text{Mi}\}}
-
\boldsymbol T_{\{\text{Pri}\}\rightarrow\{\text{Mi}\}}
\tilde{\boldsymbol p}_{\text{sem},i}^{\{\text{Pri}\}}
\right\|^{2}
},
\label{eq:sigmag_reg}
\end{equation}


\begin{figure}[!t]
\centering
\includegraphics[width=0.48\textwidth]{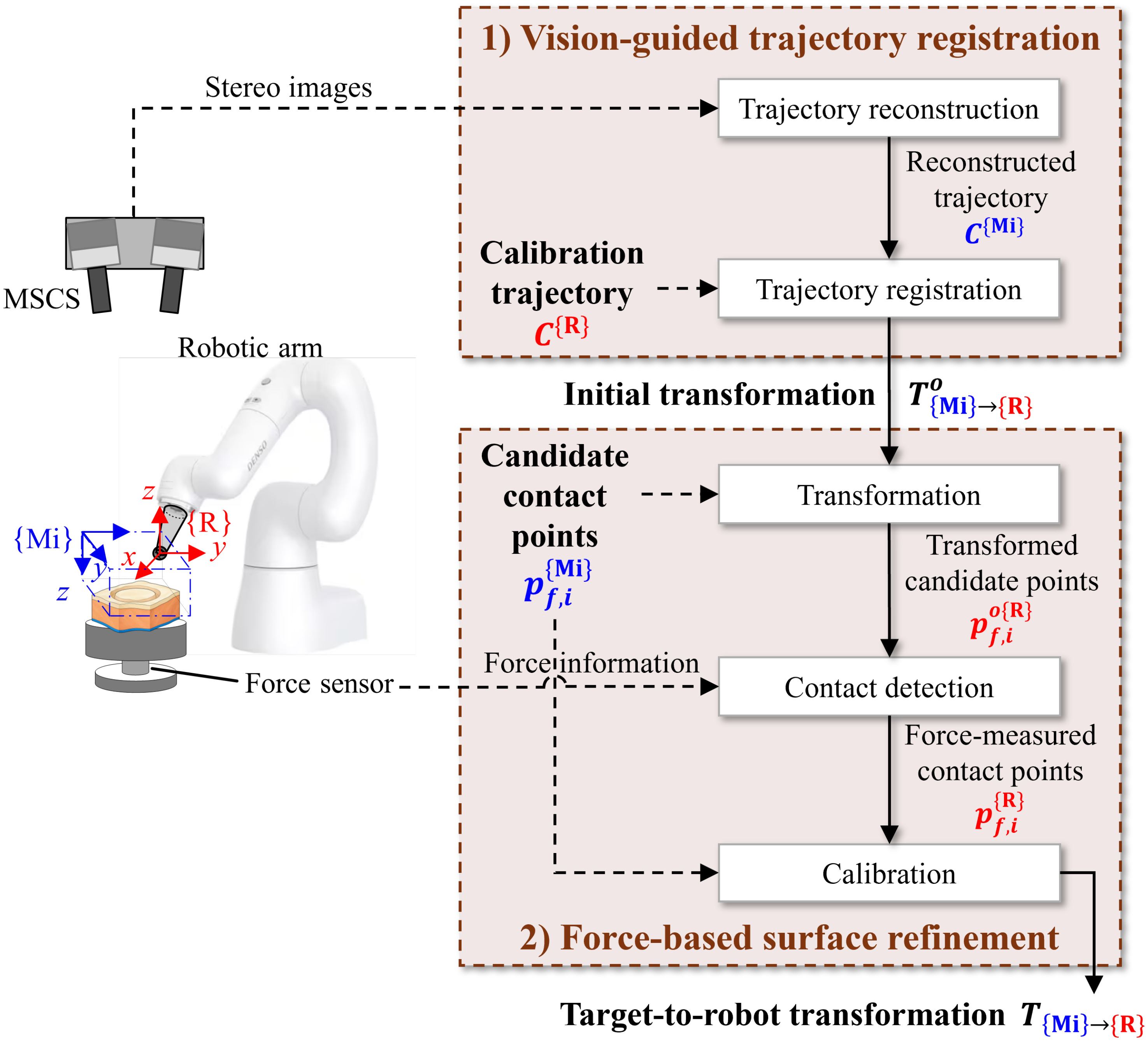}
\caption{Overview of the proposed Hybrid Vision-Force Calibration. Vision-guided trajectory registration calculates an initial transformation between $\{\text{Mi}\}$ and $\{\text{R}\}$, while force-based surface refinement introduces contact constraints to refine the transformation.}
\label{fig:calibration}
\end{figure}

\subsection{Hybrid Vision-Force Calibration}\label{Subsec:Calibration}

To transfer generic prior knowledge to robot execution, a transformation from $\{\text{Mi}\}$ to the robot coordinate frame $\{\text{R}\}$ is accurately estimated.

While MSCS provides fast and global spatial perception, its accuracy is limited, so relying exclusively on vision may leave residual discrepancies in surface alignment. Meanwhile, force sensing provides localized physical constraints through direct contact with the actual surface. Therefore, we propose a hybrid vision-force calibration strategy consisting of two stages: 
1) vision-guided trajectory registration for estimating the global transformation between $\{\text{Mi}\}$ and $\{\text{R}\}$, and 
2) force-based surface refinement for compensating remaining $\sigma_g$ using physical contact constraints, as illustrated in Fig.~\ref{fig:calibration}.

The vision-guided trajectory registration builds upon \cite{zhao2025autonomous}, which introduced a marker-free approach for microscale workspaces that estimates the transformation between $\{\text{Mi}\}$ and $\{\text{R}\}$ by aligning drill-tip positions at several discrete points. In this work, the approach is reformulated using densely sampled points along continuous trajectories. More importantly, the proposed calibration further incorporates force-based surface refinement, whose effectiveness is evaluated by an ablation study in Section~\ref{subsec:calibrationexp}.

\subsubsection{Vision-guided trajectory registration}

A calibration trajectory $\mathcal{C}^{\{\text{R}\}}$ is predefined in $\{\text{R}\}$. During execution, the MSCS continuously captures the drill-tip motion. The drill-tip locations in stereo images are detected by a CNN inspired by U-Net \cite{ronneberger2015u}, and converted to 3D positions using the calibrated stereo reconstruction model, generating the reconstructed trajectory $\mathcal{C}^{\{\text{Mi}\}}$ in $\{\text{Mi}\}$.

Given $\mathcal{C}^{\{\text{R}\}}$ and $\mathcal{C}^{\{\text{Mi}\}}$, the initial transformation $\boldsymbol T_{\{\text{Mi}\}\rightarrow\{\text{R}\}}^{o}$ is obtained using a closed-form rigid registration method.

\subsubsection{Force-based surface refinement}
\label{subsubsec:force-based}

To establish physical constraints, $N_c$ candidate contact points
$\{\boldsymbol p_{f,i}^{\{\text{Mi}\}}\}_{i=1}^{N_c}$ and their corresponding surface normals
$\{\boldsymbol n_i^{\{\text{Mi}\}}\}_{i=1}^{N_c}$ are sampled from the reconstructed target surface
$\mathcal{X}_\text{surf}$ (see Eq.~\eqref{eq:smi}).
Using the initial vision-based transformation
$\boldsymbol T_{\{\text{Mi}\}\rightarrow\{\text{R}\}}^{o}$,
the sampled points and normals are transformed into $\{\text{R}\}$ as follows:

\begin{subequations}
\begin{align}
\tilde{\boldsymbol p}_{f,i}^{o\{\text{R}\}}
&=
\boldsymbol T_{\{\text{Mi}\}\rightarrow\{\text{R}\}}^{o}
\, \tilde{\boldsymbol p}_{f,i}^{\{\text{Mi}\}}, \label{eq:trans_point} \\
\boldsymbol n_i^{\{\text{R}\}}
&=
\operatorname{Rot}
\left(
\boldsymbol T_{\{\text{Mi}\}\rightarrow\{\text{R}\}}^{o}
\right)
\boldsymbol n_i^{\{\text{Mi}\}}, \label{eq:trans_normal}
\end{align}
\end{subequations}

\noindent where $\operatorname{Rot}(\cdot)$ extracts the rotational component of the homogeneous transformation matrix.

For each $\boldsymbol p_{f,i}^{o\{\text{R}\}}$, the robot first moves to an offset point along the surface normal, and then tries to approach the surface along the negative normal direction. Physical contact is detected when the measured normal force exceeds the predefined contact threshold and remains above it for a short confirmation interval. The corresponding contact point is recorded as:

\begin{equation}
\boldsymbol p_{f,i}^{\{\text{R}\}}
=
\boldsymbol p_{f,i}^{o\{\text{R}\}} + (r_d +\Delta d_c - \Delta d_{f,i})
\boldsymbol n_i^{\{\text{R}\}},
\label{eq:pcont}
\end{equation}

\noindent where $r_d$ denotes the radius of the microdrill, $\Delta d_c$ denotes the approach offset, and $\Delta d_{f,i}$ denotes the measured displacement from the approach position to the physical contact location. The force-measured contact points are then used as physical surface constraints to refine the target-to-robot transformation:

\begin{equation}
\boldsymbol T_{\{\text{Mi}\}\rightarrow\{\text{R}\}}
=
\arg\min_{\boldsymbol T}
\sum_{i=1}^{N_c}
\left\|
\tilde{\boldsymbol p}_{f,i}^{\{\text{R}\}}
-
\boldsymbol T
\tilde{\boldsymbol p}_{f,i}^{\{\text{Mi}\}}
\right\|^{2}.
\label{eq:TmitoR}
\end{equation}

By integrating vision-based global registration with force-based surface constraints, the proposed hybrid calibration improves target-to-robot alignment, thereby enabling accurate transfer and execution of the prior-guided milling path and further reducing the calibration error $\sigma_{g,\text{cal}}$ of the geometric uncertainty $\sigma_g$:

\begin{equation}
\sigma_{g,\text{cal}}
=
\sqrt{
\frac{1}{N_c}
\sum_{i=1}^{N_c}
\left\|
\tilde{\boldsymbol p}_{f,i}^{\{\text{R}\}}
-
\boldsymbol T_{\{\text{Mi}\}\rightarrow\{\text{R}\}}
\tilde{\boldsymbol p}_{f,i}^{\{\text{Mi}\}}
\right\|^{2}
},
\label{eq:sigmag_cal}
\end{equation}

Combining the registration and calibration errors, the geometric uncertainty is given by

\begin{equation}
\sigma_g
=
\sqrt{
\sigma_{g,\text{reg}}^2
+
\sigma_{g,\text{cal}}^2
},
\label{eq:sigmag}
\end{equation}

\noindent and determines how conservatively the spatial safety margin of the milling path is chosen.

\begin{figure}[t]
\centering
\includegraphics[width=0.42\textwidth]{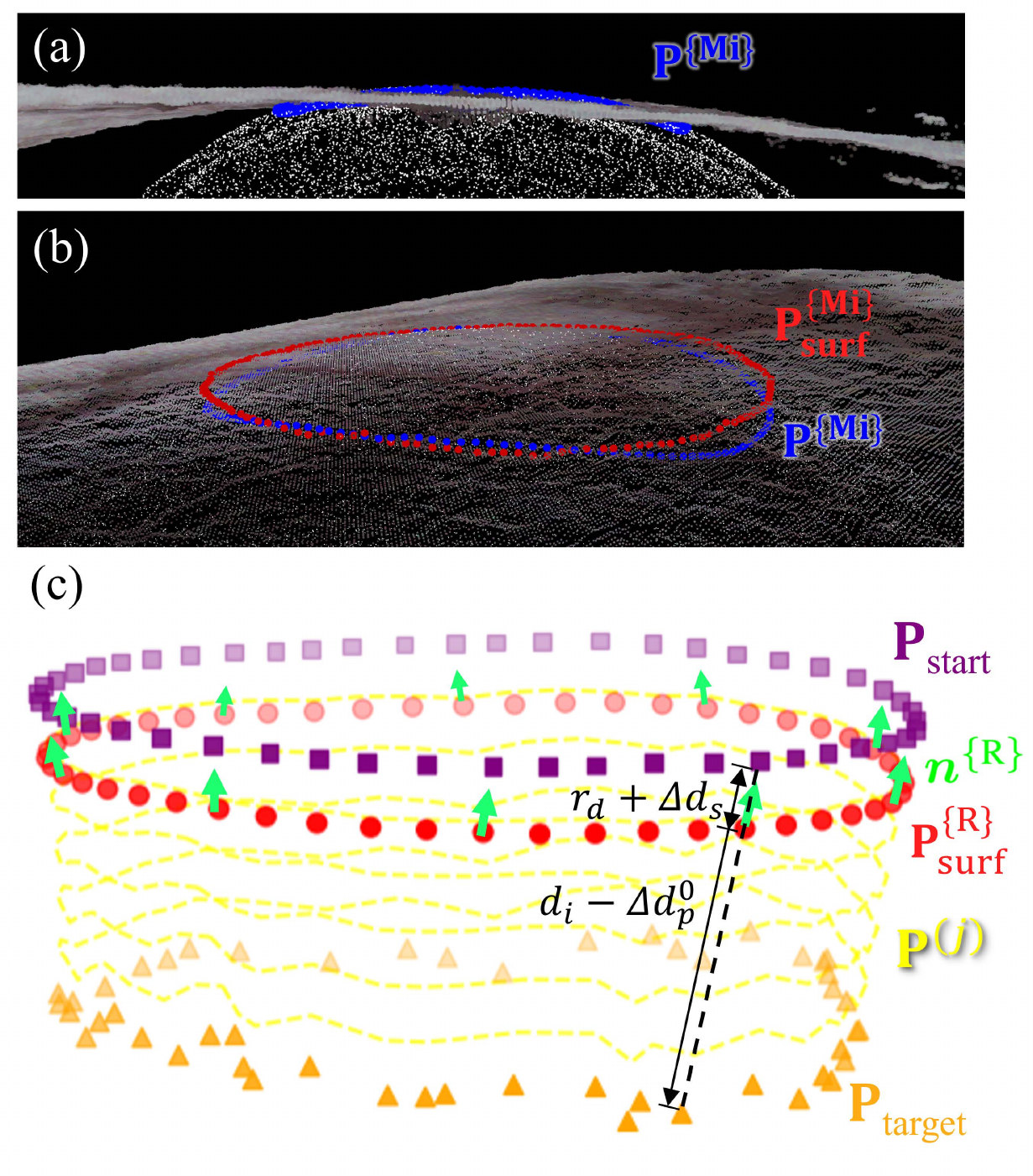}
\caption{
Generation of the prior-guided open-loop milling path set.
(a) Transformed prior-guided milling path $\mathbf{P}^{\{\text{Mi}\}}$ in $\{\text{Mi}\}$.
(b) Surface-adapted path $\mathbf{P}_{\text{surf}}^{\{\text{Mi}\}}$ obtained by projecting $\mathbf{P}^{\{\text{Mi}\}}$ to the reconstructed target surface.
(c) Interval-wise open-loop milling path generation, including the initial safe path $\mathbf{P}_{\text{start}}$, target path $\mathbf{P}_{\text{target}}$, and intermediate paths $\mathbf{P}^{1},\cdots,\mathbf{P}^{M-1}$.
}
\label{fig:openloop} 
\end{figure}

\subsection{Open-loop Milling Trajectory Generation}
\label{Subsec:pathplanner}

Although the prior-to-target and target-to-robot transformations have been obtained, the remaining geometric uncertainty $\sigma_g$ and depth uncertainty $\sigma_d$ prevent the prior-guided milling path and the thickness profile in the task-specific milling constraints $\mathcal{T}=
\left\{
\mathbf{P},
\mathbf{{D}_{P}}
\right\}$
(see Eq.~\eqref{eq:TPDP}) from being directly executed on the individual target. The path $\mathbf{P}$ is therefore further adapted to the target surface to compensate for $\sigma_g$, while the thickness profile $\mathbf{{D}_{P}}$ is used as conservative depth guidance rather than an exact measurement to address $\sigma_d$.
Finally, after transformation to $\{\text{R}\}$ and interpolation, a continuous open-loop milling trajectory with a safety margin is generated.

\subsubsection{Target surface adaptation}

First, the prior-guided milling path $\mathbf{P}
=
\{
\boldsymbol p_{i}^{\{\text{Pri}\}}
\}_{i=1}^{N}$ (see Eq.~\eqref{eq:Ppri}) is transformed into $\{\text{Mi}\}$ using the prior-to-target transformation $\boldsymbol T_{\{\text{Pri}\}\rightarrow\{\text{Mi}\}}$ (see Eq.~\eqref{eq:TpritoMi}), generating $\mathbf P^{\{\text{Mi}\}} = \{
\boldsymbol p_{i}^{\{\text{Mi}\}}
\}_{i=1}^{N}$:

\begin{equation}
\tilde{\boldsymbol p}_{i}^{\{\text{Mi}\}}
=
\boldsymbol T_{\{\text{Pri}\}\rightarrow\{\text{Mi}\}}
\tilde{\boldsymbol p}_{i}^{\{\text{Pri}\}}.
\end{equation}

As illustrated in Fig.~\ref{fig:openloop}(a) and (b), to adapt to the reconstructed target surface, $\mathbf P^{\{\text{Mi}\}}$ is projected onto $\mathcal{X}_\text{surf}$ (see Eq.~\eqref{eq:smi}) along the longitudinal direction while preserving its lateral position, resulting in the surface-adapted path $\mathbf P_{\text{surf}}^{\{\text{Mi}\}}=\{
\boldsymbol p_{\text{surf},i}^{\{\text{Mi}\}}
\}_{i=1}^{N}$.

Then, $\mathbf P_{\text{surf}}^{\{\text{Mi}\}}$ is transformed into $\{\text{R}\}$ using the target-to-robot transformation $\boldsymbol T_{\{\text{Mi}\}\rightarrow\{\text{R}\}}$ (see Eq.~\eqref{eq:TmitoR}), obtaining $\mathbf P_{\text{surf}}^{\{\text{R}\}} = \{
\boldsymbol p_{\text{surf},i}^{\{\text{R}\}}
\}_{i=1}^{N}$:

\begin{equation}
\tilde{\boldsymbol p}_{\text{surf},i}^{\{\text{R}\}}
=
\boldsymbol T_{\{\text{Mi}\}\rightarrow\{\text{R}\}}
\tilde{\boldsymbol p}_{\text{surf},i}^{\{\text{Mi}\}}.
\label{eq:PsurfR}
\end{equation}

To avoid unintended contact before milling, the starting path $\mathbf P_{\text{start}} = \{
\boldsymbol p_{\text{start},i}
\}_{i=1}^{N}$ is generated with an initial safety offset:

\begin{equation}
\boldsymbol p_{\text{start},i}
=
\boldsymbol p_{\text{surf},i}^{\{\text{R}\}}
+
(r_d+\Delta d_s)
\boldsymbol n_i^{\{\text{R}\}},
\end{equation}

\noindent
where $r_d$ denotes the radius of the microdrill, $\Delta d_s$ represents the spatial safety margin, set conservatively to bound $\sigma_g$ together with the residual reconstruction error, and $\boldsymbol n_i^{\{\text{R}\}}$ denotes the corresponding surface normal vector in $\{\text{R}\}$ (see Eq.~\eqref{eq:trans_normal}).

\subsubsection{Conservative depth milling path generation}
\label{subsubsec:depthpath}

The prior-guided thickness constraint $\mathbf{{D}_{P}}=\begin{bmatrix}
d_1,d_2,\cdots,d_N
\end{bmatrix}^\text{T}$ (see Eq.~\eqref{eq:Dp}) provides an initial estimate of the removable depth profile. Because the generic prior is population-averaged, the true thickness profile $\mathbf{{D}_{T}}=\begin{bmatrix} d_1^{\text{true}},d_2^{\text{true}},\cdots,d_N^{\text{true}} \end{bmatrix}^\text{T}$ of an individual target deviates from $\mathbf{{D}_{P}}$, so milling to the full prior depth risks overcutting thinner regions. A conservative depth margin $\Delta d_p$ is therefore reserved and scaled with the depth uncertainty $\sigma_d$, defined as the standard deviation of the element-wise deviation $\mathbf{{D}_{T}}-\mathbf{{D}_{P}}$ over the population used to construct the prior:
\begin{equation}
\sigma_d
=
\operatorname{std}\left(\mathbf{{D}_{T}}-\mathbf{{D}_{P}}\right).
\label{eq:sigmad}
\end{equation}

\noindent This margin is reserved when generating the target path $\mathbf P_{\text{target}} = \{
\boldsymbol p_{\text{target},i}
\}_{i=1}^{N}$:

\begin{equation}
\boldsymbol p_{\text{target},i}
=
\boldsymbol p_{\text{start},i}
-
(d_i-\Delta d_p)
\boldsymbol n_i^{\{\text{R}\}}.
\label{eq:ptarget}
\end{equation}

The open-loop milling paths are constructed by uniformly dividing the depth interval between $\mathbf P_{\text{start}}$ and $\mathbf P_{\text{target}}$ into $M$ segments, yielding intermediate paths $\{\mathbf P^{(j)}\}_{j=1}^{M-1}$.

A schematic of paths generation is illustrated in Fig.~\ref{fig:openloop}(c).

\subsubsection{Interpolation and trajectory generation}

The generated interval-wise milling paths are first sequentially connected according to the prescribed milling order. Constrained spline interpolation \cite{kruger2003constrained} is then applied between adjacent spatial samples to increase trajectory resolution and ensure smooth tool motion compatible with the high-precision robot controller. The resulting open-loop execution trajectory is given by

\begin{equation}
\boldsymbol{\tau}_{\text{open}}
=
\operatorname{Interp}
\left(
\mathbf{P}_{\text{start}},
\mathbf{P}^{(1)},
\cdots,
\mathbf{P}^{(M-1)},
\mathbf{P}_{\text{target}}
\right),
\label{eq:tauopen}
\end{equation}

\noindent which enables prior-guided material removal while avoiding direct penetration into the uncertain boundary. 

\section{State-Adaptive Autonomous Milling Control Architecture}
\label{Sec:state-adpative}

To address the boundary uncertainty $\sigma_b$ during execution, we propose a State-Adaptive Autonomous Milling Control Architecture, in which a finite-state machine (Section~\ref{Subsec:FSN}) coordinates open-loop milling, active boundary perception (Section~\ref{Subsec:boundaryperception}), incremental milling refinement (Section~\ref{Subsec:incremental}), and termination. This architecture extends the prior-guided open-loop execution into a perception-driven closed-loop process. Implementation details, including pseudocode of the state transition logic and mathematical derivations, are provided in the Supplementary Materials.

\subsection{Finite-State Representation of the Milling Controller}
\label{Subsec:FSN}

The autonomous milling controller is represented as a finite-state hybrid control system:

\begin{equation}
\mathcal{M}
=
\left\{
\mathcal{S},
\mathcal{G},
\delta,
S_0
\right\},
\label{eq:MSG}
\end{equation}

\noindent where $\mathcal{S}$ denotes the set of operational states (introduced in Section~\ref{subsubsec:staterepresentation}), $\mathcal{G}$ denotes the set of guard conditions governing state transitions (introduced in Section~\ref{Subsec:Statetransition}), $\delta$ denotes the state transition function, and $S_0$ represents the initial state.

The state transition function is generally represented as

\begin{equation}
S_{k+1}
=
\delta(S_k,\boldsymbol{x}_k),
\end{equation}

\noindent where $\boldsymbol{x}_k$ denotes the operational condition at the $k$-th execution step that governs state transitions:

\begin{equation}
\boldsymbol{x}_k = (\boldsymbol{d}_k, \Gamma_{\text{trans},k}, \sigma_{b,k}),
\end{equation}

\noindent where $\boldsymbol{d}_k$ denotes the cumulative milling depth profile along the milling path at the $k$-th execution step, $\Gamma_{\text{trans},k}\in[0,1]$ denotes the global boundary transition indicator of the remaining structure, representing the mean boundary-transition status over all probing locations (formally defined by Eq.~\eqref{eq:gammatrans}), and $\sigma_{b,k}$ denotes the boundary uncertainty, characterized by the spatial consistency of the local boundary-transition responses, as formally defined in Eq.~\eqref{eq:sigmab}.

\subsubsection{Definition of operational states}\label{subsubsec:staterepresentation}

The operational states set $\mathcal{S}$ is defined as

\begin{equation}
\mathcal{S}
=
\left\{
S_{\text{open}},
S_{\text{perc}},
S_{\text{ref}},
S_{\text{term}}
\right\},
\end{equation}

\noindent where the individual states are described as follows.

\begin{itemize}

\item \textbf{Open-loop milling state} $S_{\text{open}}$:
This is the initial state of the controller ($S_0=S_{\text{open}}$). The robot activates the microdrill rotation and executes the prior-guided trajectory $\boldsymbol{\tau}_{\text{open}}$ (defined by Eq.~\eqref{eq:tauopen}) to remove the majority of the material based on the estimated thickness, maintaining a conservative margin above the predicted boundary.

\item \textbf{Active boundary perception state} $S_{\text{perc}}$:
The milling is suspended, and the robot performs force-based probing at multiple locations using the non-rotating microdrill to assess the global detachability of the remaining structure. This state is detailed in Section~\ref{Subsec:boundaryperception}.

\item \textbf{Incremental refinement state} $S_{\text{ref}}$: 
The controller updates the local depth of the refinement trajectory according to the estimated local boundary status. The robot reactivates the microdrill and performs incremental milling refinement. This state is detailed in Section~\ref{Subsec:incremental}.

\item \textbf{Termination state} $S_{\text{term}}$:
The autonomous milling procedure is successfully concluded.

\end{itemize}

\begin{figure}[t]
\centering
\includegraphics[width=0.45\textwidth]{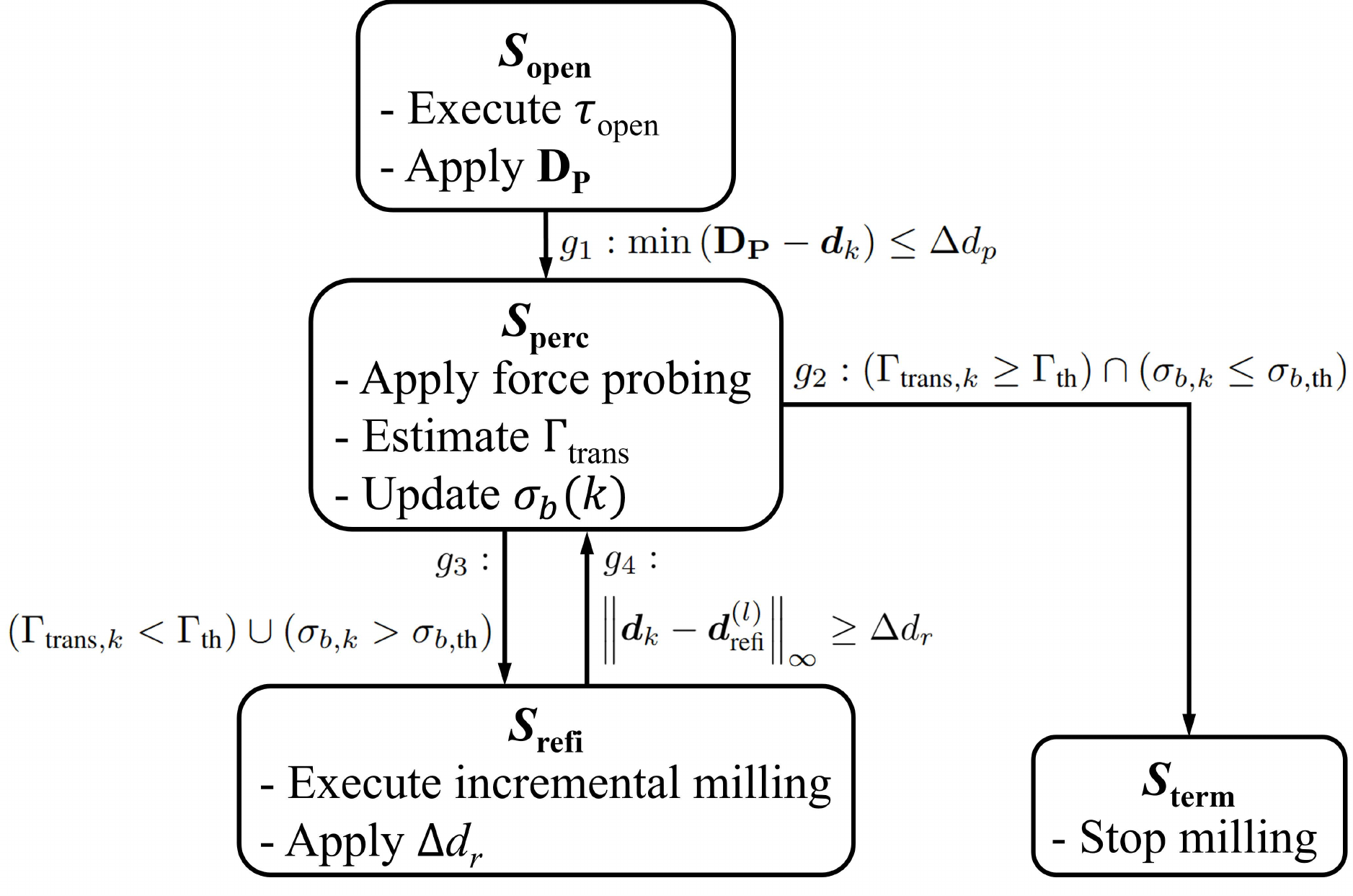}
\caption{Finite-state machine of the proposed state-adaptive autonomous milling framework.}
\label{fig:FSM}
\end{figure}

\subsubsection{State transition strategy}
\label{Subsec:Statetransition}

The transitions between operational states are governed by the guard-condition set $\mathcal{G}$ (Fig.~\ref{fig:FSM}), which is defined as 

\begin{equation}
\mathcal{G}
=
\left\{
g_1,
g_2,
g_3,
g_4
\right\}.
\end{equation}

Initially, the controller starts from $S_{\text{open}}$ and executes $\boldsymbol{\tau}_{\text{open}}$. It transfers to $S_{\text{perc}}$ when the remaining structural thickness reaches the conservative depth margin $\Delta d_p$ defined in Eq.~\eqref{eq:ptarget}:

\begin{equation}
g_1:
\min \left( \mathbf{{D}_P} - \boldsymbol{d}_k \right)
\leq
\Delta d_p,
\label{eq:g1}
\end{equation}

\noindent where $\mathbf{{D}_P}$ is the prior-guided thickness profile (see Eq.~\eqref{eq:Dp}). The $\min(\cdot)$ operator triggers the transition once any local remaining thickness falls below the margin.

During $S_{\text{perc}}$, the controller enters $S_{\text{term}}$ when both the boundary transition condition and the boundary-uncertainty requirement for safe termination are satisfied:

\begin{equation}
g_2:
\left(
\Gamma_{\text{trans},k}
\geq
\Gamma_{\text{th}}
\right)
\cap
\left(
\sigma_{b,k}
\leq
\sigma_{b,\text{th}}
\right),
\label{eq:g2}
\end{equation}

\noindent where $\Gamma_{\text{th}}$ is the predefined completion threshold, and $\sigma_{b,\text{th}}$ is the maximum allowable boundary uncertainty for autonomous termination. Otherwise, the controller enters $S_{\text{ref}}$ for incremental material removal:

\begin{equation}
g_3:
\left(
\Gamma_{\text{trans},k}
<
\Gamma_{\text{th}}
\right)
\cup
\left(
\sigma_{b,k}
>
\sigma_{b,\text{th}}
\right).
\label{eq:g3}
\end{equation}

During $S_{\text{ref}}$, the refinement depth is bounded by a conservative increment $\Delta d_r$. Assuming that the current $S_{\text{ref}}$ is the $l$-th refinement iteration, the controller returns to $S_{\text{perc}}$ after $\Delta d_r$ is reached:

\begin{equation}
g_4:
\left\|
\boldsymbol{d}_k
-
\boldsymbol{d}_{\text{ref}}^{(l)}
\right\|_{\infty}
\geq
\Delta d_r,
\label{eq:g4}
\end{equation}

\noindent where $\boldsymbol{d}_{\text{ref}}^{(l)}$ denotes the cumulative material removal profile recorded at the beginning of the $l$-th refinement iteration. The infinity norm $\Vert \cdot \Vert_{\infty}$ triggers the transition once any local depth increment reaches $\Delta d_r$ during the non-uniform milling introduced in Section~\ref{Subsec:incremental}.

The perception-refinement cycle therefore repeats until $g_2$ is satisfied.

\subsection{Active Boundary Perception}
\label{Subsec:boundaryperception}

\begin{figure}[t]
\centering
\includegraphics[width=3.1in]{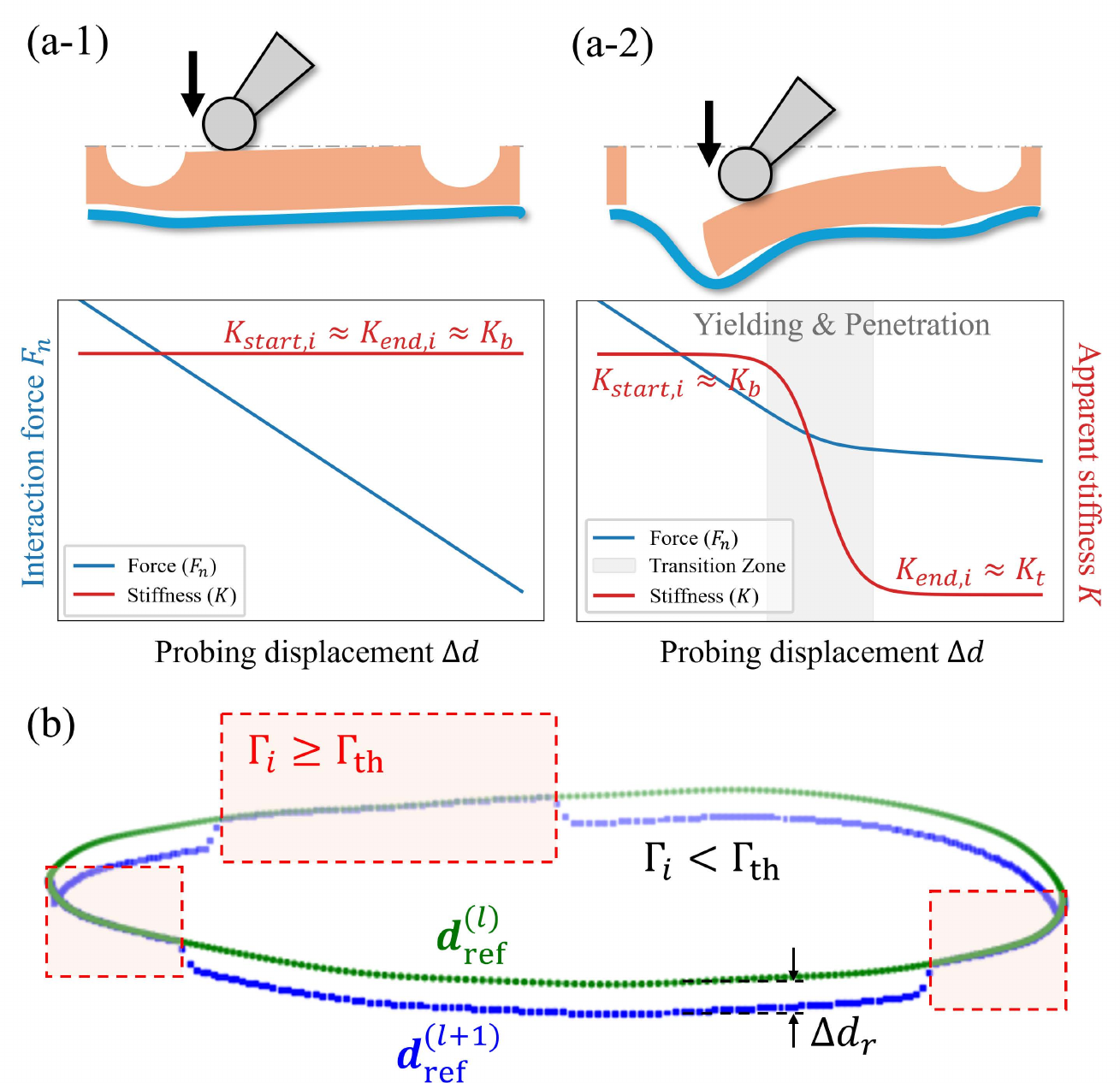}
\caption{
(a) Schematic and force-displacement response of active probing at two local boundary statuses: (a-1) an intact boundary and (a-2) a penetrated boundary. (b) Schematic of the local boundary status-based milling depth update in incremental refinement. The depth profiles $\boldsymbol{d}_{\text{ref}}^{(l)}$ and $\boldsymbol{d}_{\text{ref}}^{(l+1)}$ represent the milling depth before and after the $l$-th refinement iteration, respectively. Regions identified as penetrated (within red box) retain their previous depth, whereas intact regions receive an incremental depth update $\Delta d_r$.
}
\label{fig:5-palpation-loc-strategy}
\end{figure}

After entering $S_{\text{perc}}$, the controller stops the microdrill rotation and performs controlled probing at predefined locations. The measured force-displacement responses are characterized in terms of relative stiffness to identify local changes in mechanical support, estimate boundary status during execution, and assess the global detachability of the remaining structure.

The active boundary perception strategy builds upon the probing-based approach in \cite{lin2025object}, which estimates absolute local mechanical stiffness and applies a hard threshold for binary local penetration decisions followed by a strict voting rule for global detachability. In contrast, the proposed method utilizes a continuous local support-loss index derived from relative stiffness changes, reducing dependence on structure-specific absolute stiffness. By aggregating these continuous local indices, the framework provides more informative local boundary estimation and a more robust structural-level assessment. Its effectiveness is demonstrated through a direct comparison with \cite{lin2025object} on the same biological surrogate, as described in Section~\ref{subsec:exp3}.

\subsubsection{Boundary probing points generation}

The active boundary perception does not evaluate every point along the milling contour. Instead, a limited number $N_p$ of representative probing locations $\{\boldsymbol p_{\text{bp},i}^{\{\text R\}}\}_{i=1}^{N_p}$ are selected from the planned surface path and shifted inward into the remaining structure. This strategy avoids contact with the machined groove and enables reliable force-based estimation of local boundary status. 

\subsubsection{Controlled probing interaction}

Similar to the contact detection procedure in Section~\ref{subsubsec:force-based}, the robot physically contacts each probing point $\boldsymbol p^{\{\text R\}}_{\text{bp},i}$ and identifies the actual physical contact position $\boldsymbol{p}^{\{\text{R}\}}_{\text{con},i}$ (see Eqs.~\eqref{eq:trans_normal} and \eqref{eq:pcont}).

Subsequently, the robot performs a controlled probing motion along the negative surface normal direction:

\begin{equation}
\boldsymbol{p}^{\{\text{R}\}}_i(t)
=
\boldsymbol{p}^{\{\text{R}\}}_{\text{con},i}
-
\Delta d_i(t)
\boldsymbol{n}^{\{\text{R}\}}_i,
\quad \text{with } 0 \leq \Delta d_i(t) \leq d_m,
\end{equation}

\noindent where $\Delta d_i(t)$ represents the additional inward displacement after surface contact is established, and $d_m$ denotes a preset maximum safety displacement that provides sufficient structural response for boundary estimation while maintaining a safe interaction range for the underlying fragile tissue.

\subsubsection{Force-displacement response analysis}

During probing, the axial force response is continuously recorded as a function of displacement, and the apparent stiffness of the contacted structure is estimated from the local force-displacement slope:

\begin{equation}
K_i(\Delta d_i)
=
\left|
\frac{\partial f(\Delta d_i)}
{\partial \Delta d_i}
\right|.
\end{equation}

As illustrated in Fig.~\ref{fig:5-palpation-loc-strategy}(a-1), when the local boundary is intact, the probing response is dominated by the rigid support of the remaining structure. Therefore, the apparent stiffness remains approximately constant, with
$K_{\text{start},i} \approx K_{\text{end},i} \approx k_b$,
where $k_b$ denotes the apparent stiffness associated with the intact boundary.

Conversely, as illustrated in Fig.~\ref{fig:5-palpation-loc-strategy}(a-2), when the local boundary loses mechanical support and becomes penetrated, the probing response transitions to a lower apparent stiffness, with
$K_{\text{start},i} \approx k_b$ and $K_{\text{end},i} \approx k_t$, where $k_t < k_b$.

These distinct stiffness responses motivate a normalized support-loss index:

\begin{equation}
\Gamma_i = \operatorname{clip} \left( 1 - \frac{K_{\text{end},i}}{K_{\text{start},i}}, 0, 1 \right),
\label{eq:itai}
\end{equation}

\noindent where the $\operatorname{clip}(\cdot, 0, 1)$ function strictly bounds the output between 0 and 1 to mitigate the effects of sensory noise, ensuring a consistent physical interpretation. Accordingly, an intact boundary yields $\Gamma_i\approx0$, whereas a pronounced loss of mechanical support produces a larger $\Gamma_i$. Unlike structure-specific force thresholds, $\Gamma_i$ characterizes the relative change in mechanical support, enabling boundary status estimation across different biological structures.

\subsubsection{Detachability identification}

To estimate the global detachability of the remaining structure, the local support-loss indices from multiple probing locations are aggregated:

\begin{equation}
\Gamma_{\text{trans}}
=
\frac{1}{N_p}
\sum_{i=1}^{N_p}
\Gamma_i .
\label{eq:gammatrans}
\end{equation}

\noindent
where $\Gamma_{\text{trans}}$ represents the global boundary transition indicator of the structure. A larger value suggests a transition toward a detachable structure.

Furthermore, the boundary uncertainty $\sigma_b$ is characterized by the spatial consistency of the measured boundary-transition responses, computed as the spatial dispersion of the local support-loss indices:

\begin{equation}
\sigma_b
=
\sqrt{
\frac{1}{N_p - 1}
\sum_{i=1}^{N_p}
(\Gamma_i-\Gamma_{\text{trans}})^2
+\sigma_f^2
},
\label{eq:sigmab}
\end{equation}

\noindent where the variance term characterizes the spatial inconsistency among the probing locations, and $\sigma_f$ accounts for a constant noise floor identified from sensor calibration. The resulting $\sigma_b$, together with $\Gamma_{\text{trans}}$, is used by the finite-state controller to evaluate boundary completion and determine termination or refinement according to $g_2$ and $g_3$ (see Eqs.~\eqref{eq:g2} and \eqref{eq:g3}).

\subsection{Incremental Milling Refinement}
\label{Subsec:incremental}

Once the controller switches to $S_{\text{ref}}$, rather than applying a uniform additional depth along the entire contour, the proposed strategy performs spatially adaptive material removal based on local boundary status.

Specifically, the milling depth is updated according to the local support-loss index $\Gamma_i$ (see Eq.~\eqref{eq:itai}), as illustrated in Fig.~\ref{fig:5-palpation-loc-strategy}(b). The discrete probing locations divide the milling path into corresponding segments, and the milling depth for each segment is updated by:

\begin{equation}
d_{\text{ref}, i}^{(l+1)}
=
\begin{cases}
d_{\text{ref}, i}^{(l)}+\Delta d_r, & \Gamma_i<\Gamma_{\text{th}},\\
d_{\text{ref}, i}^{(l)}, & \Gamma_i\geq\Gamma_{\text{th}},
\end{cases}
\label{eq:refinement_update}
\end{equation}

\noindent
where $d_{\text{ref}, i}^{(l)}$ denotes the $i$-th element of the cumulative material removal profile $\boldsymbol{d}_{\text{ref}}^{(l)}$ recorded at the beginning of the $l$-th refinement iteration, and $\Delta d_r$ is the conservative increment, scaled in relation to $\sigma_d$, to limit material removal under the remaining boundary uncertainty $\sigma_{b,k}$.

The robot then resumes milling along the original lateral contour using the updated depth profile $\boldsymbol{d}_{\text{ref}}^{(l+1)}$. To ensure kinematic continuity without overcutting, the profile is smoothed via a Hanning-window low-pass filter, while depths in already penetrated regions are strictly clamped to their previously executed values.

Once $g_4$ is satisfied (see Eq.~\eqref{eq:g4}), the controller transitions back to $S_{\text{perc}}$ for subsequent boundary perception. The probing locations $\{\boldsymbol p_{\text{bp},i}^{\{\text R\}}\}$ are updated according to the previous local boundary status, excluding penetrated regions ($\Gamma_i \geq \Gamma_{\text{th}}$) and re-evaluating only intact regions ($\Gamma_i < \Gamma_{\text{th}}$) in the following perception-refinement cycle. The $\Gamma_i$ values of excluded penetrated regions are retained for subsequent computation of $\Gamma_{\text{trans}}$ and $\sigma_b$ according to Eqs.~\eqref{eq:gammatrans} and \eqref{eq:sigmab}.

This iterative perception-refinement process progressively reduces the remaining unresolved regions toward the desired boundary status without relying on aggressive one-step material removal. The procedure continues until the completion criterion $g_2$ is satisfied (see Eq.~\eqref{eq:g2}), after which the controller enters $S_{\text{term}}$.

\section{Experiment}\label{Sec:experiment}

To evaluate the proposed autonomous precision biological milling framework, a series of hierarchical experiments was conducted using biological surrogates and in vivo mouse models.

Experiment~I evaluated anatomy-aware prior-to-target registration and hybrid vision-force calibration. Experiment~II characterized the baseline milling execution capability through lateral trajectory realization and material removal consistency. Experiment~III evaluated active boundary perception for classifying local boundary status and global detachability. Experiment~IV evaluated the integrated state-adaptive framework through circular window creation and non-circular milling tasks. Finally, Experiment~V provided in vivo validation in a real biological environment.

All experiments used the robotic platform in Section~\ref{Sec:hard}. The controller parameters governing active boundary perception and state transitions were calibrated using an independent set of biological surrogate samples and remained unchanged throughout Experiments~III--V, without being retuned for individual specimens. Detailed calibration protocol and other experimental details, including hardware and software configurations, parameter settings, metric formulations, and trial-level analysis, are provided in the Supplementary Materials.

\subsection{Biological Surrogates and Animal Preparation}
\label{Sec:experiment_setup}

Chicken eggs were employed as surrogates in Experiments II--IV. The eggshell reproduces several challenges associated with biological precision milling, including uneven surface curvature, spatial variation in thin-shell thickness ($\approx 0.32~\text{mm}$), and a fragile inner membrane boundary. These characteristics enable systematic evaluation of machining precision, active boundary perception, and adaptive milling control before in vivo validation.

Mouse cranial window creation was selected as the final in vivo validation task (Experiment~V) because it not only exhibits the task-relevant uncertainties of general precision biological milling, as discussed in Section~\ref{Sec:introduction}, but also demands high execution accuracy: millimeter-scale lateral positioning and sub-millimeter material removal for windows of varying sizes (2--6~mm) and geometries, while avoiding damage to the underlying boundary. Together, these characteristics provide a representative validation scenario for the proposed framework.

For mouse cranial window creation, female NOD.CB17-Prkdc\textsuperscript{scid}/J mice aged 6--8 weeks and weighing 20--25~g were used. All procedures were approved by the Animal Care and Use Committee, Graduate School of Medicine, The University of Tokyo (Approval No.: A2023M042-07). Under anesthesia, the surface of the mouse skull was exposed, and the head was rigidly fixed using a stereotaxic head holder to minimize head motion during experiments.

\begin{figure}[t]
\centering
\includegraphics[width=\columnwidth]{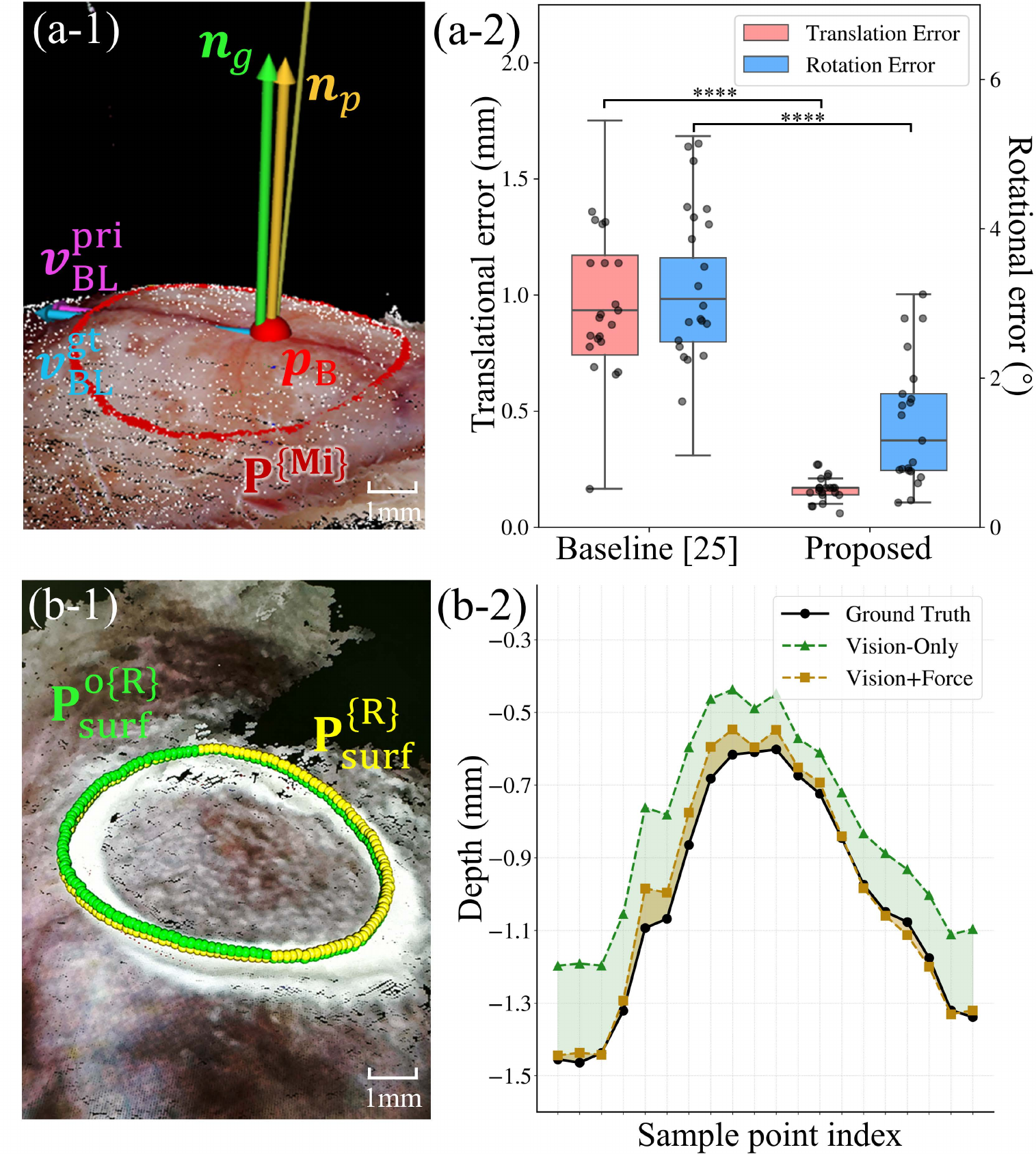}
\caption{
Results of Experiment I, including evaluations of proposed (a) prior-to-target registration and (b) target-to-robot calibration.
(a-1) Visualization of transformed Bregma position $\boldsymbol{p}_{\text{B}}$, anatomical orientation vectors $\boldsymbol v_{\text{BL}}^{\text{gt}}$ and $\boldsymbol v_{\text{BL}}^{\text{pri}}$, corresponding surface normals $\boldsymbol n_g$ and $\boldsymbol n_p$, and the transformed prior-guided path $\mathbf P^{\{\text{Mi}\}}$. (a-2) Box plots summarizing the translational and rotational errors for the baseline \cite{lin2024} and the proposed method, with outliers excluded.
(b-1) Visualization of transformed paths before and after force-based refinement, including $\mathbf{P}_{\text{surf}}^{o\{\text{R}\}}$ and $\mathbf{P}_{\text{surf}}^{\{\text{R}\}}$. (b-2) Point-wise comparison of estimated depth before and after force-based refinement at each sample point with the ground truth.
}
\label{fig:ex1_Result}
\end{figure}

\subsection{Experiment I: Validation of Generic Anatomical Prior-Based Planning} \label{subsec:exp1}

This experiment evaluated the proposed Generic Anatomical Prior-Based Planning (Section~\ref{Sec:preop}) in reducing the geometric uncertainty, assessing anatomy-aware prior-to-target registration followed by hybrid vision-force calibration.

\begin{figure*}[t]
\centering
\includegraphics[width=0.88\textwidth]{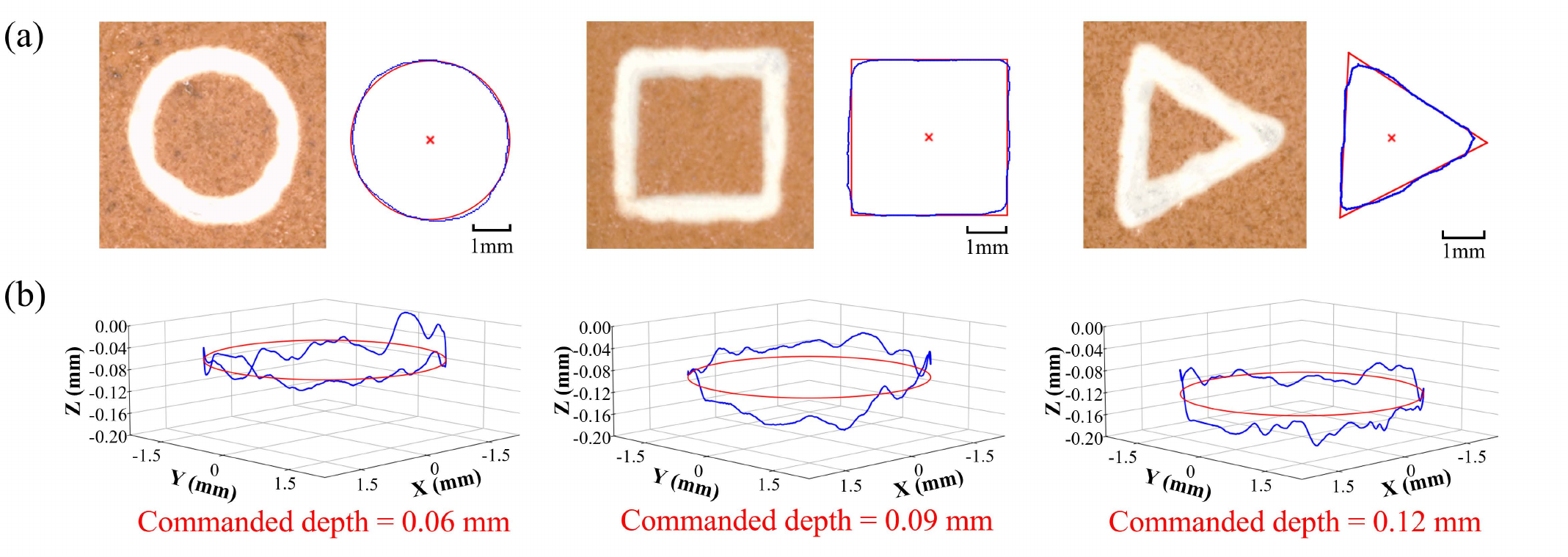}
\caption{
Representative results of Experiment~II.
(a) Lateral contour evaluation of different milling patterns. Red and blue curves denote the reference and reconstructed milling contours, respectively.
(b) Material removal depth evaluation under different commanded depths. Red and blue curves denote the commanded and achieved depths, respectively.
}\label{fig:exp2_result} 
\end{figure*}

\subsubsection{Evaluation of anatomy-aware prior-to-target registration}
\label{subsubsec:registrationexp}

To evaluate the prior-to-target registration described in Section~\ref{Subsec:alignment}, we registered the generic prior \cite{maga2017population} to 21 reconstructed mouse cranial surfaces. The evaluation was conducted at two levels: an anatomical-level comparison against the baseline method \cite{lin2024}, and an independent task-level assessment using metrics specific to our milling framework.

\paragraph{Anatomical-level comparison}

\begin{table}[t]
\centering
\caption{Comparison of anatomical-level registration accuracy}
\label{tab:registration_results}
\begin{tabular}{llcc}
\toprule
\textbf{Method} 
& \textbf{Metric} 
& \textbf{Mean $\pm$ Std. Dev.} 
& \textbf{Max} 
\\
\midrule
\multirow{2}{*}{Baseline \cite{lin2024}} 
& Translational error (mm)          
& 0.990 $\pm$ 0.535  
& 5.103  
\\
& Rotational error ($^\circ$) 
& 3.28 $\pm$ 2.52  
& 26.34  
\\
\midrule
\multirow{2}{*}{\textbf{Proposed}} 
& Translational error (mm)          
& \textbf{0.164 $\pm$ 0.055} 
& \textbf{0.270} 
\\
& Rotational error ($^\circ$) 
& \textbf{1.40 $\pm$ 0.85} 
& \textbf{3.12} 
\\
\bottomrule
\end{tabular}
\end{table}

At the anatomical level, the registration accuracy was quantified against manually annotated ground truth using two metrics: (1) translational error (Euclidean distance between the reference and registered Bregma positions $\boldsymbol{p}_{\text{B}}$), (2) rotational error (angular deviation between transferred and reference Bregma-Lambda vectors $\boldsymbol v_{\text{BL}}^{\text{gt}}$ and $\boldsymbol v_{\text{BL}}^{\text{pri}}$), as illustrated in Fig.~\ref{fig:ex1_Result}(a-1).

Fig.~\ref{fig:ex1_Result}(a-2) and Table~\ref{tab:registration_results} summarize the quantitative anatomical-level registration results. Compared with the baseline \cite{lin2024}, the proposed method achieved an 83.4\% reduction in mean translational error and a 57.3\% reduction in mean rotational error (both $p<0.0001$, paired t-test). These improvements demonstrate that the redesigned anatomy-aware registration establishes an accurate and robust correspondence between the generic prior and individual targets.

\paragraph{Task-level assessment}

Beyond anatomical metrics, task-level registration accuracy was evaluated using two trajectory-oriented indicators: (1) surface normal error (angular deviation between corresponding surface normals $\boldsymbol{n}_g$ and $\boldsymbol{n}_p$), and (2) path transfer error (lateral deviation in $N=160$ points between a reference path defined on the reconstructed target surface and $\mathbf{P}^{\{\text{Mi}\}}$, the transformed prior-guided milling path before projection), as illustrated in Fig.~\ref{fig:ex1_Result}(a).

The proposed method achieved a mean surface normal error of $9.31\pm 4.15^\circ$ (Max: $14.27^\circ$). Although this angular deviation is relatively large, it primarily reflects the remaining geometric uncertainty after registration arising from inherent inter-subject variations in local surface morphology rather than inaccurate anatomical correspondence. Importantly, this deviation has limited impact on path transfer, as the prior-guided path is not directly used to generate the final milling trajectory but is instead adapted to the individual target through surface projection after registration, as described in Section~\ref{Subsec:pathplanner}. This is further supported by the path transfer error, which was only $0.074\pm0.032$~mm (Max: $0.212$~mm). These results demonstrate that, despite the remaining geometric uncertainty, the registered generic prior provides reliable global positioning and can be effectively adapted to the individual target anatomy.

\subsubsection{Ablation study of hybrid vision-force calibration}\label{subsec:calibrationexp}

To evaluate the contribution of force-based surface refinement, an ablation study is conducted by comparing vision-only calibration with the proposed hybrid vision-force calibration. 

Following the proposed hybrid calibration framework introduced in Section~\ref{Subsec:Calibration}, the target-surface-adapted path $\mathbf{P}_{\text{surf}}^{\{\text{Mi}\}}$ was transformed into $\{\text{R}\}$ using the initial vision-based transformation $\boldsymbol{T}^{o}_{\{\text{Mi}\}\rightarrow\{\text{R}\}}$ and the refined transformation $\boldsymbol{T}_{\{\text{Mi}\}\rightarrow\{\text{R}\}}$ (obtained from $N_c=8$ independent force contact landmarks), generating $\mathbf{P}_{\text{surf}}^{o\{\text{R}\}}$ and $\mathbf{P}_{\text{surf}}^{\{\text{R}\}}$, as illustrated in Fig.~\ref{fig:ex1_Result}(b-1). Twenty points with the same path indices were sampled from each path and physically contacted on the actual surface to obtain the ground truth depth at each point. The surface alignment was evaluated using the depth error between each sampled path point and its corresponding ground truth.

As shown in Fig.~\ref{fig:ex1_Result}(b-2) and Table~\ref{tab:surface_calibration}, compared with the vision-only baseline, force-based refinement reduced the alignment error at all corresponding point indices. The Mean Absolute Error (MAE) and RMSE across all positions were also significantly reduced after refinement ($p<0.0001$, paired t-test), demonstrating improved depth alignment and facilitating subsequent trajectory generation and autonomous milling.

\begin{figure*}[t]
\centering
\includegraphics[width=\textwidth]{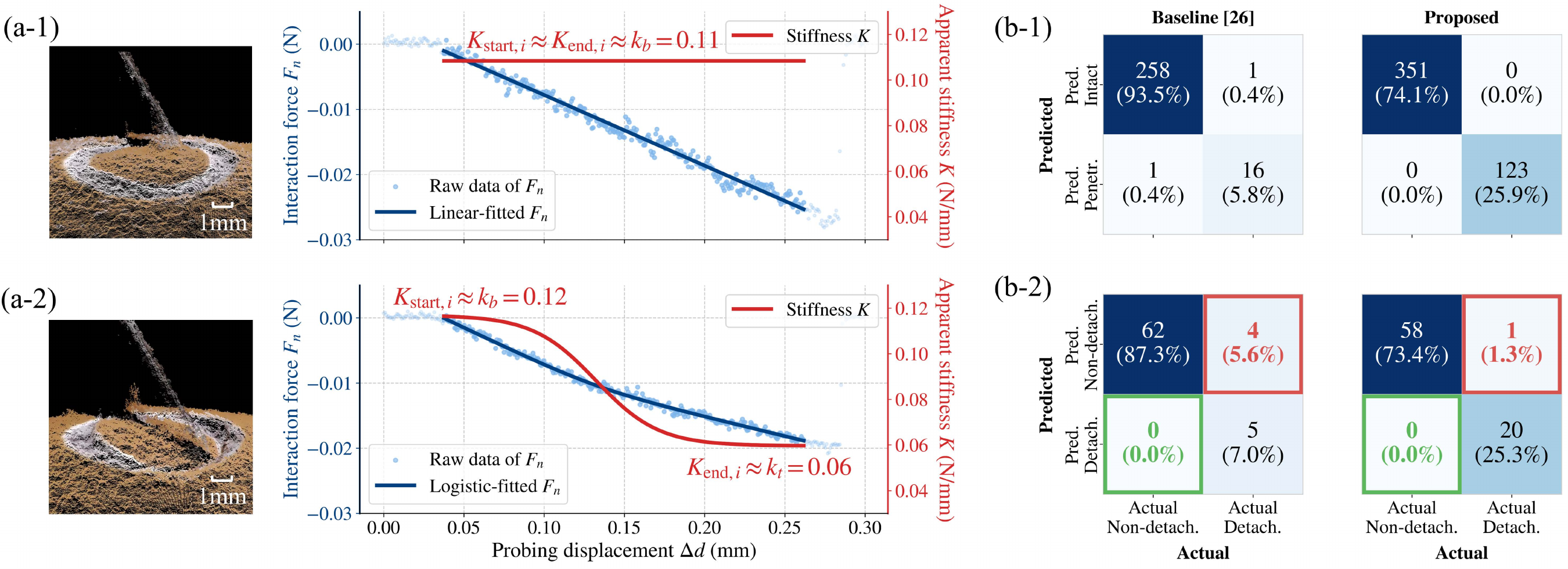}
\caption{Results of Experiment~III.
(a) Representative MSCS stereo-vision point clouds and corresponding force-displacement responses for (a-1) intact and (a-2) penetrated local boundary status, respectively.
(b) Confusion matrices for (b-1) local boundary-condition classification and (b-2) global detachability classification, with the baseline \cite{lin2025object} and proposed methods shown on the left and right, respectively. In (b-2), red and green boxes highlight false negative (FN) and false positive (FP) classifications, respectively.}
\label{fig:perception_confusion}
\end{figure*}

\begin{table}[t]
\centering
\caption{Comparison of surface alignment accuracy}
\label{tab:surface_calibration}
\begin{tabular}{ccc}
\toprule
\textbf{Method} & \textbf{MAE (mm)} & \textbf{RMSE (mm)} \\
\midrule
Baseline (vision only) & 0.198 $\pm$ 0.068 & 0.210 \\
\textbf{Proposed} & \textbf{0.037 $\pm$ 0.031} & \textbf{0.048} \\
\bottomrule
\end{tabular}
\end{table}

\subsection{Experiment II: Quantitative Evaluation of Robotic Milling Geometric Accuracy}
\label{subsec:exp2}

In this experiment, we evaluated the task-level geometric accuracy of open-loop robotic milling under biological interaction conditions. Rather than characterizing the intrinsic motion accuracy of the robotic manipulator, the evaluation focused on how accurately the intended geometries and material removal depths were achieved on the target. Therefore, geometric accuracy was assessed from the post-operative surface morphology reconstructed by a surface profilometer, rather than the nominal tool trajectory itself. Lateral accuracy was quantified from the machined $x$-$y$ contours, while axial consistency was assessed from the material removal depth profiles. Each pattern and depth was repeated three times on independent samples.

\begin{table}[t]
\centering
\caption{Lateral contour accuracy across repeated path executions}
\label{tab:trajectory_execution}
\setlength{\tabcolsep}{4pt}
\begin{tabular}{lccc}
\toprule
\multirow{2}{*}{\textbf{Pattern}} & \textbf{Mean Error} & \textbf{Max Error} & \textbf{Std. Dev.} \\
                                  & \textbf{(mm)}       & \textbf{(mm)}      & \textbf{(mm)} \\
\midrule
Circle ($r=2~\text{mm}$)   & 0.037 & 0.117 & 0.028 \\
Circle ($r=3~\text{mm}$)   & 0.040  & 0.149  & 0.029 \\
Circle ($r=4~\text{mm}$)   & 0.043 & 0.176  & 0.037 \\
Square ($L=4~\text{mm}$)   & 0.039  & 0.128  & 0.035  \\
Triangle ($L=4~\text{mm}$) & 0.042  & 0.182  & 0.032 \\
\bottomrule
\end{tabular}
\end{table}

\subsubsection{Lateral contour accuracy}

To evaluate lateral contour accuracy, diverse milling patterns were fabricated on curved eggshell surfaces, including circles ($r=2,\ 3,\ 4$~mm), squares ($L=4$~mm), and equilateral triangles ($L=4$~mm). The machined contours were registered to their corresponding ideal templates, and point-wise contour deviations were calculated for quantitative analysis, as illustrated in Fig.~\ref{fig:exp2_result}(a).

As shown in Table~\ref{tab:trajectory_execution}, all patterns achieved a mean contour error below $0.050$~mm, demonstrating sufficient lateral fidelity for microscale biological milling. Circular trajectories exhibited relatively uniform deviations. In contrast, larger local deviations (up to $0.182$~mm) occurred near sharp corners (e.g., the triangle), which we attribute to the finite tool diameter interacting with the sharp commanded path ($r_d=0.5$~mm). This localized distortion is acceptable for precision milling of biological structures, where overall geometric fidelity is prioritized over exact reproduction of sharp geometric features.

\subsubsection{Material removal depth consistency}

To assess axial material removal consistency, circular milling patterns of $r=2$~mm were fabricated at commanded removal depths of $0.06$, $0.09$, and $0.12$~mm. Representative depth profiles comparing the actual material removal depths with the commanded depths along the milling paths are shown in Fig.~\ref{fig:exp2_result}(b).

As summarized in Table~\ref{tab:depth_consistency}, the mean depth errors remained small ($<0.010$~mm), indicating that the average achieved depth was close to the commanded value. However, the relatively large RMSE values ($\approx 0.026$~mm) revealed marked point-wise variations between the commanded and achieved depths along the milling path. These variations were attributed to local discrepancies between the predefined removal depth and the actual material response: different locations along the biological surface differ in local curvature, thickness, and structural stiffness. During milling, these variations alter the contact condition, interaction force, and deformation of the structure under microdrill loading, resulting in spatially varying material removal depths that cannot be accurately compensated by predefined open-loop commands.

Therefore, although open-loop control can achieve accurate lateral trajectory execution, it cannot guarantee consistent axial material removal under heterogeneous biological conditions. This limitation motivates the proposed state-adaptive closed-loop framework, where active boundary perception identifies the actual structural boundary and compensates for local variations during execution.

\begin{table}[t]
\centering
\caption{Material removal consistency at different commanded depths}
\label{tab:depth_consistency}
\begin{tabular}{ccc}
\toprule
\textbf{Commanded Depth (mm)} & \textbf{Mean Error (mm)} & \textbf{RMSE (mm)} \\
\midrule
0.06 & 0.000 & 0.027 \\
0.09 & 0.005 & 0.027 \\
0.12 & 0.008 & 0.025 \\
\bottomrule
\end{tabular}
\end{table}

\begin{table}[t]
\centering
\caption{Quantitative Evaluation of Active Boundary Perception}
\label{tab:perception_results}
\setlength{\tabcolsep}{3.5pt}
\begin{tabular}{llcccc}
\toprule
\textbf{Method}
& \textbf{Level}
& \textbf{Acc. (\%)}
& \textbf{Prec. (\%)}
& \textbf{Rec. (\%)}
& \textbf{F1 Score} \\
\midrule
\multirow{2}{*}{Baseline \cite{lin2025object}}
& Local
& 99.28 & 94.12 & 94.12 & 0.941 \\
& Global
& 94.37 & 100.00 & 55.56 & 0.714 \\
\midrule
\multirow{2}{*}{\textbf{Proposed}}
& Local
& \textbf{100.00} & \textbf{100.00} & \textbf{100.00} & \textbf{1.000} \\
& Global
& \textbf{98.73} & \textbf{100.00} & \textbf{95.24} & \textbf{0.976} \\
\bottomrule
\end{tabular}
\end{table}

\subsection{Experiment III: Evaluation of Active Boundary Perception}
\label{subsec:exp3}

In this experiment, the proposed active boundary perception introduced in Section~\ref{Subsec:boundaryperception} is evaluated in comparison with the baseline probing method of \cite{lin2025object}.

\subsubsection{Experimental design and evaluation metrics}

The baseline and proposed methods were evaluated on 12 and 20 independent eggshells, respectively. Perception was evaluated at two levels: local boundary status at individual probing locations, classified as intact or penetrated, and global detachability of the remaining structure, classified as detachable or non-detachable. With $N_p=4$ probing locations for the baseline and $N_p=6$ for the proposed method, 276 and 474 local classification results and 71 and 79 global classification results were obtained, respectively. Ground truth labels were manually determined from MSCS stereo-vision observations of local deformation and structural detachability.

\subsubsection{Results and analysis}

Representative observations of intact and penetrated boundary locations are shown in Fig.~\ref{fig:perception_confusion}(a). Quantitative classification results are summarized in Table~\ref{tab:perception_results}, with corresponding confusion matrices shown in Fig.~\ref{fig:perception_confusion}(b).

At the local level, the proposed method achieved 100\% accuracy (vs.~99.28\%) with zero misclassifications. Unlike the absolute-stiffness baseline, the relative-stiffness-based $\Gamma_i$ reduces sensitivity to inter-sample variations, providing a more discriminative measure of local boundary status.

At the global level, the aggregation of the continuous local support-loss indices $\Gamma_i$, represented by $\Gamma_{\text{trans}}$, improved recall from 55.56\% to 95.24\%, reducing false negatives from four to one while maintaining 100\% precision. By contrast, the baseline's binary voting discards the magnitude information of local responses. The sole remaining false negative occurred when near-threshold local responses kept $\Gamma_{\text{trans}}$ below the detachability threshold, resulting in a conservative decision.

Overall, the proposed method improved both local classification and global detachability recall under the evaluated configurations, with near-threshold responses remaining a limitation of the current aggregation strategy.

\subsection{Experiment IV: State-Adaptive Autonomous Window Creation}
\label{subsec:exp4}

As Experiment~II revealed the limitations of open-loop depth control under biological conditions, this experiment evaluated the proposed state-adaptive framework for autonomous window creation on biological surrogates. A quantitative comparison between the proposed closed-loop framework and the baseline open-loop strategy is first conducted, followed by an evaluation of the framework's shape adaptability through the creation of complex, non-circular windows.

\subsubsection{Comparison of open-loop and proposed strategies}
\label{subsubsec:exp4_comparison}

For comparison, 40 eggs were randomly divided into an open-loop control group and a proposed closed-loop group ($n=20$ each) to create circular windows with $r=3~\text{mm}$.

\begin{figure}[t]
\centering
\includegraphics[width=0.42\textwidth]{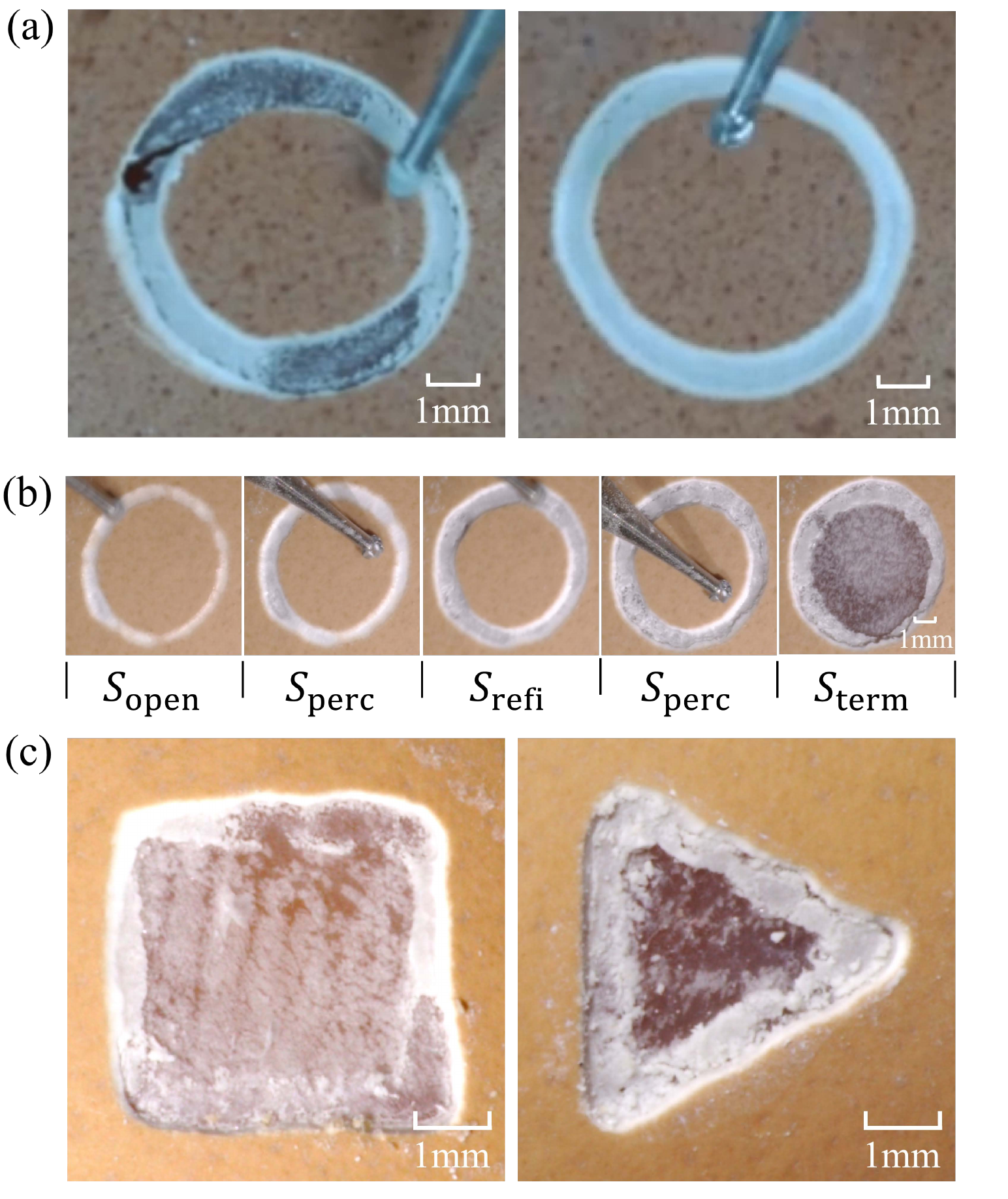}
\caption{
Results of Experiment IV.
(a) Representative failure modes of open-loop milling, including membrane damage and incomplete milling.
(b) Autonomous closed-loop milling process and corresponding operational states.
(c) Autonomous window creation of other geometries, including square and triangular patterns.}
\label{fig:exp3_result} 
\end{figure}

\begin{table}[t]
\centering
\caption{Comparison of Autonomous Window Creation Performance}
\label{tab:window_creation_results}
\setlength{\tabcolsep}{3pt}
\begin{tabular}{lccccc}
\toprule
\multirow{2}{*}{\textbf{Method}} & \textbf{Success} & \textbf{Membrane} & \textbf{Incomplete} & \textbf{Refine.} & \textbf{Time} \\
                                 & \textbf{Rate}    & \textbf{Damage}   & \textbf{Milling}    & \textbf{Iters.}  & \textbf{(min)} \\
\midrule
Open-loop      & 25\%             & 30\%              & 45\%                & N/A              & 16.5 \\
\textbf{Proposed}       & \textbf{100\%}            & \textbf{0\%}               & \textbf{0\%}                 & 2.95           & 23.8 \\
\bottomrule
\end{tabular}
\end{table}

\paragraph{Experimental design and evaluation metrics}

The open-loop group executed a fixed milling depth based on the statistical average shell thickness ($0.32~\text{mm}$), whereas the proposed group employed the closed-loop adaptive strategy.

Outcomes were classified into three categories: 

\begin{itemize}
    \item \textbf{Success:} complete window creation without damaging the inner membrane.
    \item \textbf{Membrane damage:} rupture caused by excessive penetration.
    \item \textbf{Incomplete milling:} insufficient material removal preventing window completion.
\end{itemize}

Moreover, the execution time and refinement iterations were recorded to compare the robustness and efficiency.

\paragraph{Results and analysis}

Quantitative results are summarized in Table~\ref{tab:window_creation_results}. The open-loop strategy achieved a success rate of only 25\%, as the fixed-depth command could not compensate for local thickness variations, resulting in either membrane damage (30\% of trials) or incomplete milling (45\% of trials), as shown in Fig.~\ref{fig:exp3_result}(a).

In contrast, the proposed framework achieved a 100\% success rate. This improvement was enabled by autonomous state adaptation, where the robot dynamically transitioned among milling, active perception, refinement, and termination states based on online information, as illustrated in Fig.~\ref{fig:exp3_result}(b). Through iterative perception and local refinement, the milling trajectory was progressively adapted to the actual structural boundary, selectively refining intact regions while avoiding unnecessary material removal in penetrated regions.

The iterative refinement process required an average of 2.95 iterations per trial, increasing the average execution time from 16.5 to 23.8~min. The additional time primarily resulted from the conservative probing parameters used to prevent membrane damage, reflecting an intentional efficiency-safety trade-off that prioritizes reliable boundary adaptation under complex biological conditions.

\subsubsection{Autonomous window creation of different geometries}

To evaluate the generalization capability of the proposed state-adaptive control framework beyond circular paths, square and equilateral triangular windows with $L=4$~mm ($n=5$ per shape) were autonomously created on eggshells without modifying the control architecture.

All polygonal windows were successfully created, and representative results are shown in Fig.~\ref{fig:exp3_result}(c). Despite the abrupt directional transitions and local variations in tool-material interaction near sharp corners, the framework preserved the intended window geometries without damaging the underlying membrane. These results demonstrate the generalization of the proposed control strategy to diverse milling geometries.

\begin{figure*}[t]
\centering
\includegraphics[width=\textwidth]{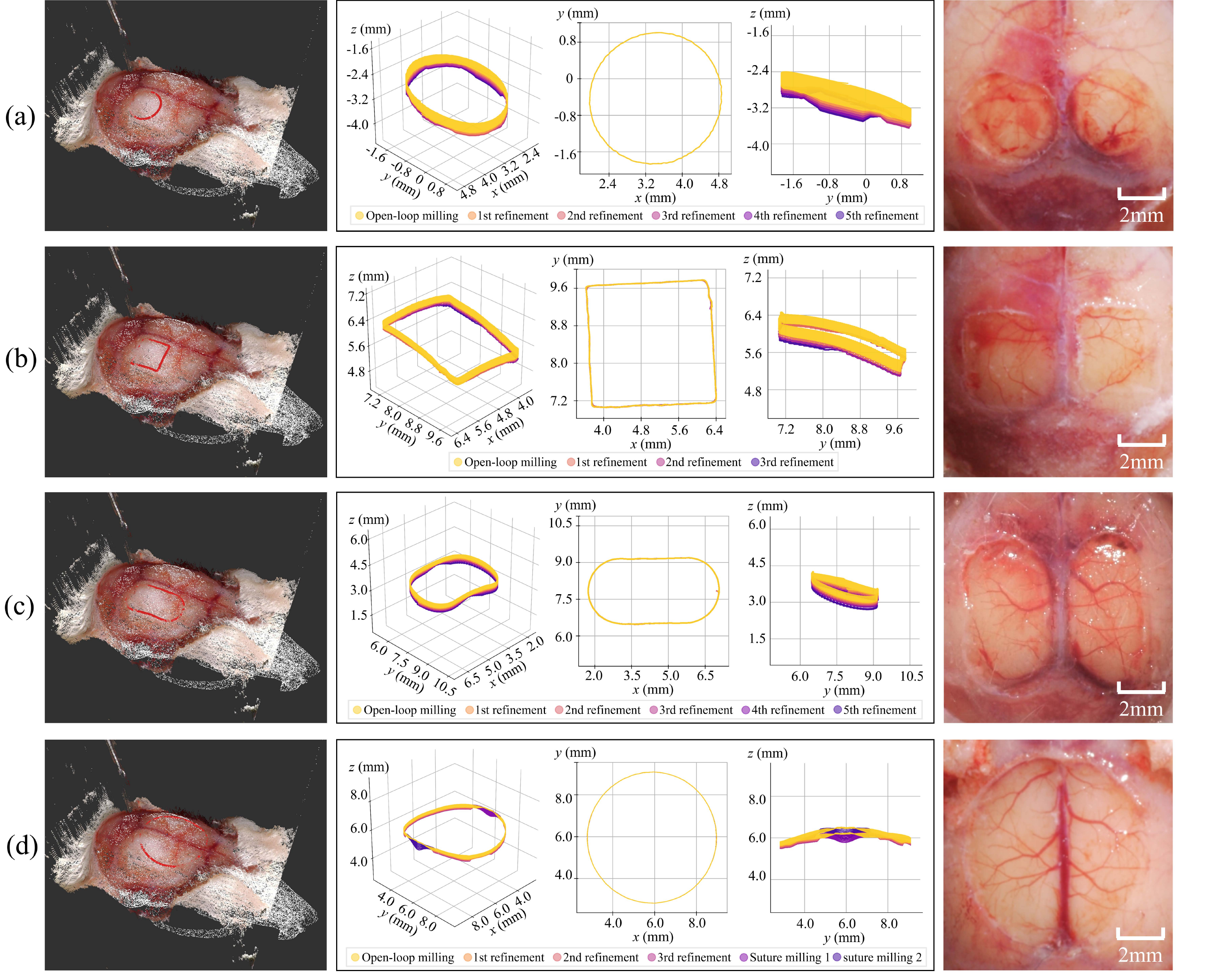}
\caption{
Representative results of autonomous cranial window creation for (a) circular, (b) square, (c) obround, and (d) large cross-suture circular windows, showing the prior-guided milling path registered to the target surface, complete robot trajectory, and milling outcome. The 3D, top, and side views of the trajectories illustrate the window contours, and the material removal during the initial open-loop milling and successive refinement.
}\label{fig:exp5_traj} 
\end{figure*}

\subsection{Experiment V: Autonomous Cranial Window Creation in Real Biological Environment}
\label{subsec:exp5}

This experiment evaluated the proposed framework in a real biological environment through autonomous in vivo mouse cranial window creation.

\subsubsection{Experimental design and evaluation metrics}

To evaluate the generalization capability of the proposed autonomous milling framework, 15 in vivo trials were conducted across four representative cranial window geometries: circular, obround, square, and large cross-suture circular. The number of trials for each geometry was determined by experimental availability and ethical considerations rather than for statistical comparison among geometries.

For large cross-suture circular windows, a dedicated milling procedure was adopted, differing from the trajectory generation and depth-update strategy used for small windows to account for the highly vascularized suture regions. The suture locations were identified from the anatomical prior, and the tool depth was maintained at a constant level when passing through these regions, while the remaining regions were milled using the standard trajectory generation and depth control until all non-suture regions were penetrated. Afterward, the suture regions were milled separately using a localized depth-controlled strategy to ensure complete bone-flap detachment while minimizing bleeding.

A trial was considered successful when the planned window geometry was completely created while preserving the integrity of the underlying biological structures and avoiding unintended damage to the surrounding skull.

\subsubsection{Results and analysis}

All 15 trials were successfully completed, demonstrating the feasibility of the proposed framework for autonomous precision milling in real biological environments. Representative results are shown in Fig.~\ref{fig:exp5_traj}. Superficial skull bleeding occurred in 10 of the 15 trials and was managed through brief hemostatic pauses. No dural injury was observed. Such events were therefore not considered failures, and all mice recovered uneventfully after the procedure. Importantly, the superficial bleeding did not interfere with the boundary perception or refinement process, and the temporary pauses were introduced solely for animal welfare rather than to compensate for algorithmic limitations.

The execution time ranged from 8.0 to 46.5 min, depending on the window geometry. In particular, large cross-suture circular windows required longer execution times because of the larger milling areas, additional refinement iterations, and separate milling of the suture regions. The separate suture milling also required more frequent hemostatic pauses to manage bleeding. Consequently, the combination of iterative refinement and safety-driven pauses increased overall execution time compared with previously reported open-loop approaches \cite{ghanbari2019, navabi25, li2025optical}. This reflects an intentional trade-off between efficiency and safety in precision biological milling.

Overall, these results demonstrate that the proposed framework can be transferred from surrogate models to real biological environments, accommodating heterogeneous anatomical structures while maintaining reliable boundary adaptation.

\section{Discussion}
\label{Sec:discussion}

The experimental results support a complementary division of roles between generic anatomical priors and active boundary perception. Semantic-guided registration and hybrid vision-force calibration transfer population-level anatomy to individual targets, providing conservative global guidance for milling. Active probing then estimates local support loss and global detachability, addressing boundary status that a static prior cannot predict. The framework does not require explicit probabilistic inference: geometric uncertainty is accommodated by a safety margin informed by registration and calibration errors (Tables~\ref{tab:registration_results} and~\ref{tab:surface_calibration}), depth uncertainty by a conservative margin scaled to the population-derived thickness variation $\sigma_d$, and boundary uncertainty by $\sigma_b$ together with $\Gamma_{\text{trans}}$, which jointly determine termination or further refinement.

Several limitations remain. First, the current framework assumes a relatively stable target configuration and does not explicitly compensate for dynamic deformation or physiological motion. Second, conservative probing and incremental refinement improve boundary reliability but increase execution time, which may be reduced through learned force-response models and adaptive refinement policies. Finally, superficial bleeding and other biological events are not explicitly modeled and required hemostatic pauses in the in vivo experiments. Future work will investigate dynamic tracking, deformation estimation, and event-aware handling to improve robustness in more diverse biological environments.

The present study does not establish clinical reliability or safety. Nevertheless, the proposed framework provides a general strategy for autonomous precision milling of biological structures by combining generic anatomical priors, active boundary perception, and task-level state adaptation.

\section{Conclusion}

This article presented a state-adaptive framework for autonomous precision milling of biological structures under incomplete anatomical knowledge. The framework combines generic anatomical priors, transferred to individual targets through semantic-guided registration and hybrid vision-force calibration, with active boundary perception and incremental refinement to adapt milling to actual structural conditions. Experiments on biological surrogates demonstrated a 100\% success rate in autonomous window creation, compared with 25\% for fixed-depth open-loop milling, while all 15 in vivo mouse trials were completed without observed dural injury. These results demonstrate the feasibility of combining population-level anatomical knowledge with online physical interaction for autonomous precision milling in biological environments.

\bibliographystyle{IEEEtran}
\bibliography{references}

\begin{IEEEbiography}[{\includegraphics[width=1in,height=1.25in,clip]{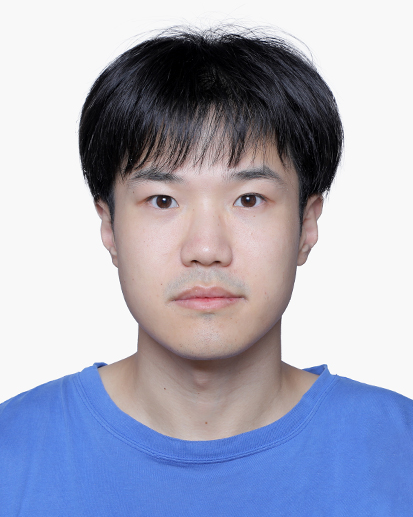}}]{Enduo Zhao} received the B.S. degree in mechanical engineering from Tsinghua University, Beijing, China, in 2018, and the M.Sc. and Ph.D. degrees in mechanical engineering from The University of Tokyo, Tokyo, Japan, in 2020 and 2025, respectively. He is currently a Post-Doctoral Researcher with the School of Biomedical Engineering, Tsinghua University. His research interests include automation, surgical robotics, and medical image processing.
\end{IEEEbiography}%

\begin{IEEEbiography}[{\includegraphics[width=1in,height=1.25in,clip]{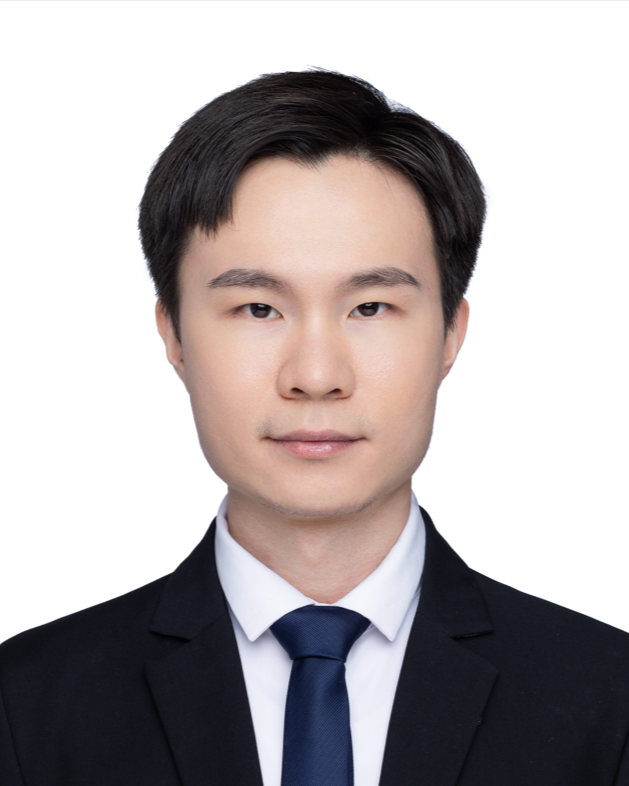}}]{Xiaofeng Lin} received his B.E. degree from Huazhong University of Science and Technology in 2018, M.E. degree from University of Chinese Academy of Sciences in 2021, and Ph.D. degree from the University of Tokyo in 2024. He is a post-doctoral researcher with Graduate School of Medicine, the University of Tokyo. He is interested in surgical navigation, medical robots, and 3D Vision. 
\end{IEEEbiography}%

\begin{IEEEbiography}[{\includegraphics[width=1in,height=1.25in,clip]{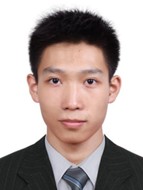}}]{Yifan Wang} received his B.E. degree in mechanical engineering from Xi'an Jiaotong University, Xi'an, China, in 2020, and the M.E. and Ph.D. degrees in mechanical engineering from The University of Tokyo, Tokyo,Japan, in 2023 and 2026, respectively. His research interests include artificial intelligence, surgical robotics and computer vision.
\end{IEEEbiography}%

\begin{IEEEbiography}[{\includegraphics[width=1in,height=1.25in,clip]{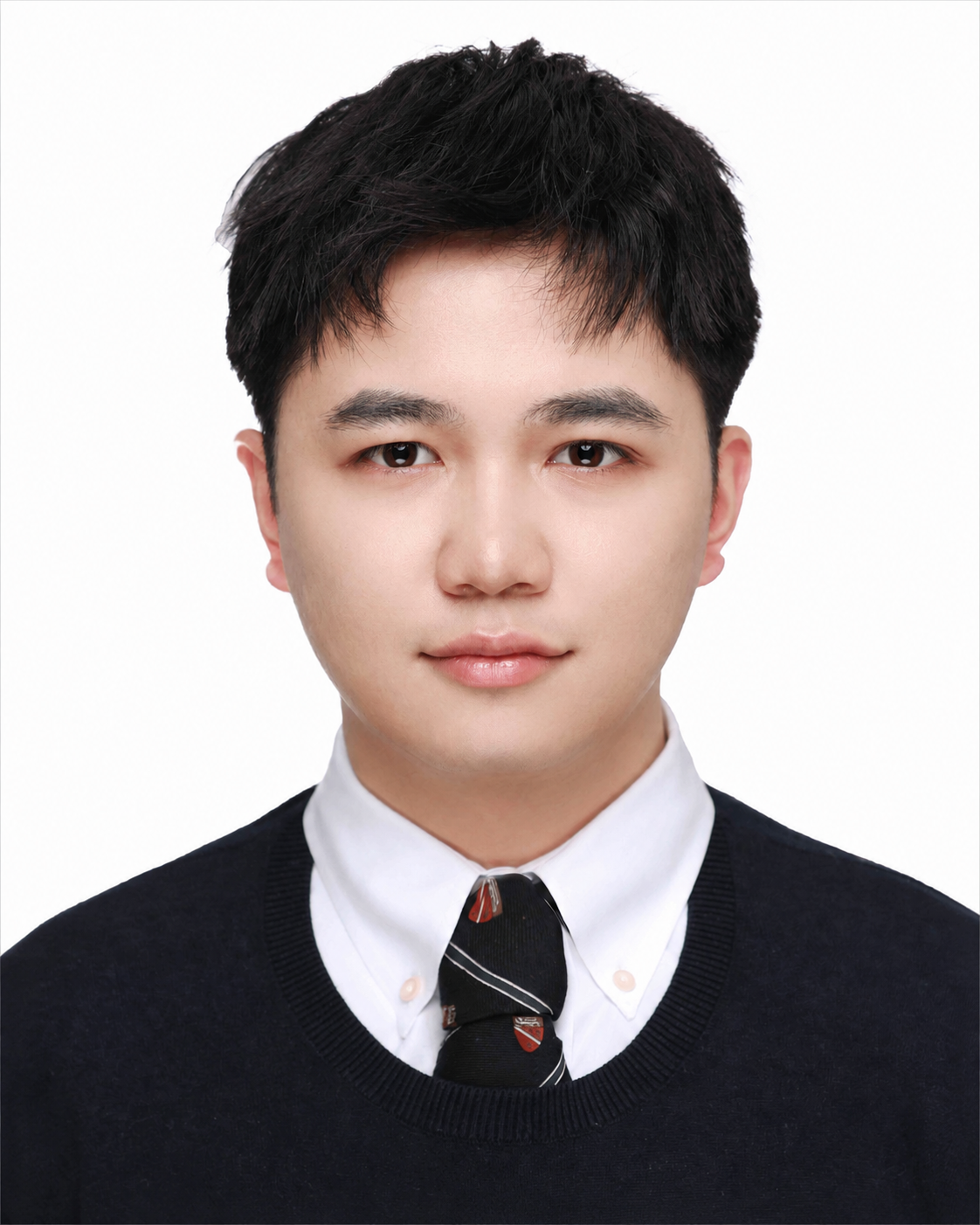}}]{Yuhan Song} received the B.Eng. degree in information security from Hefei University of Technology, Hefei, China, in 2019, and the M.Sc. degree from the Japan Advanced Institute of Science and Technology, Ishikawa, Japan, in 2023. He is currently pursuing the Ph.D. degree with The University of Tokyo, Tokyo, Japan, where he is also a JSPS Research Fellow. His research interests include computer vision, medical artificial intelligence, and intelligent robotics.
\end{IEEEbiography}%

\begin{IEEEbiography}[{\includegraphics[width=1in,height=1.25in,clip]{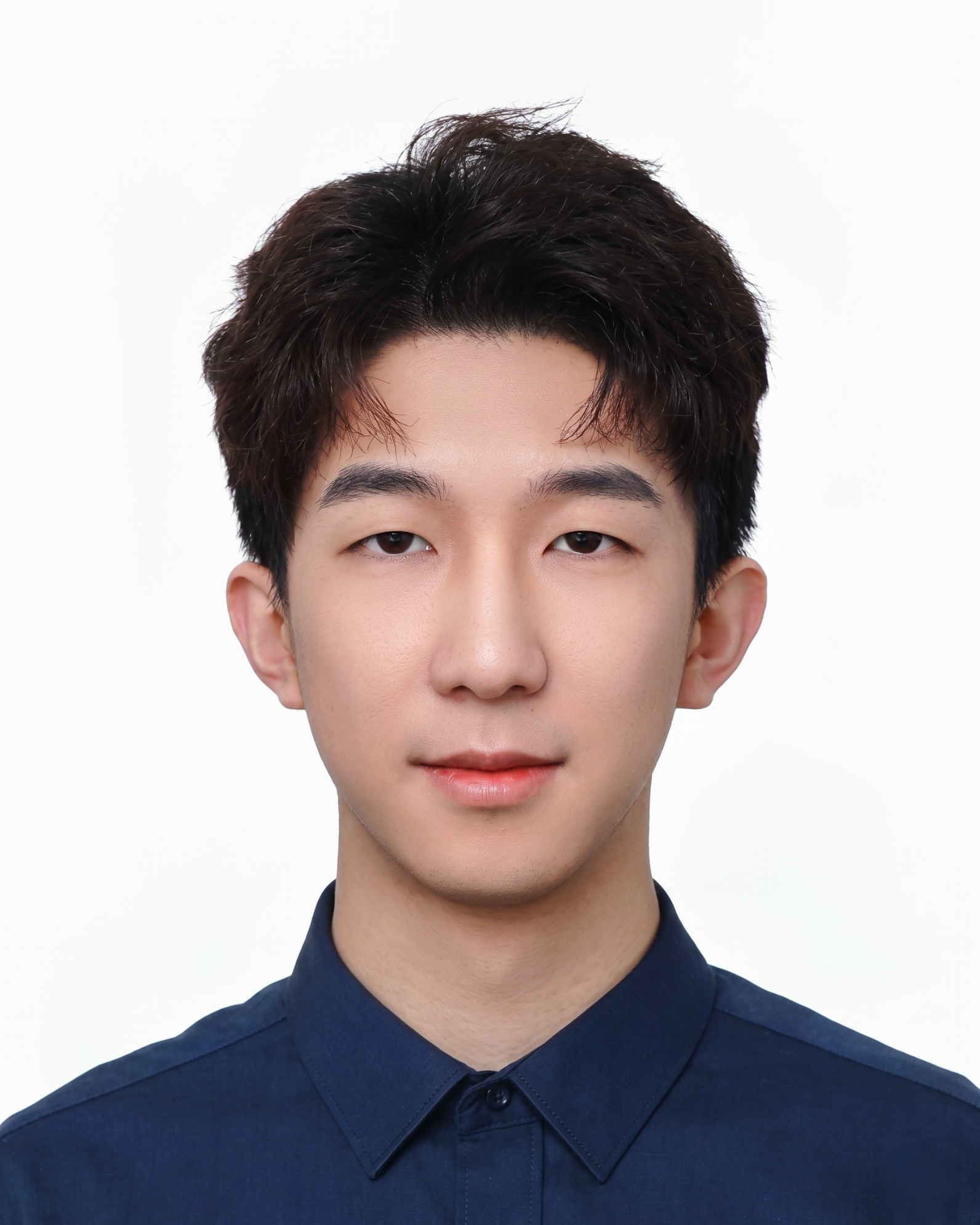}}]{Weihan Li} received the B.S. degree in automation from Harbin Institute of Technology, Shenzhen, China, in 2023. He is currently working toward the M.Sc. degree in mechanical engineering with The University of Tokyo, Tokyo, Japan. His research interests include robotics, AI4Science, and scientific discovery.    
\end{IEEEbiography}%

\begin{IEEEbiography}[{\includegraphics[width=1in,height=1.25in,clip]{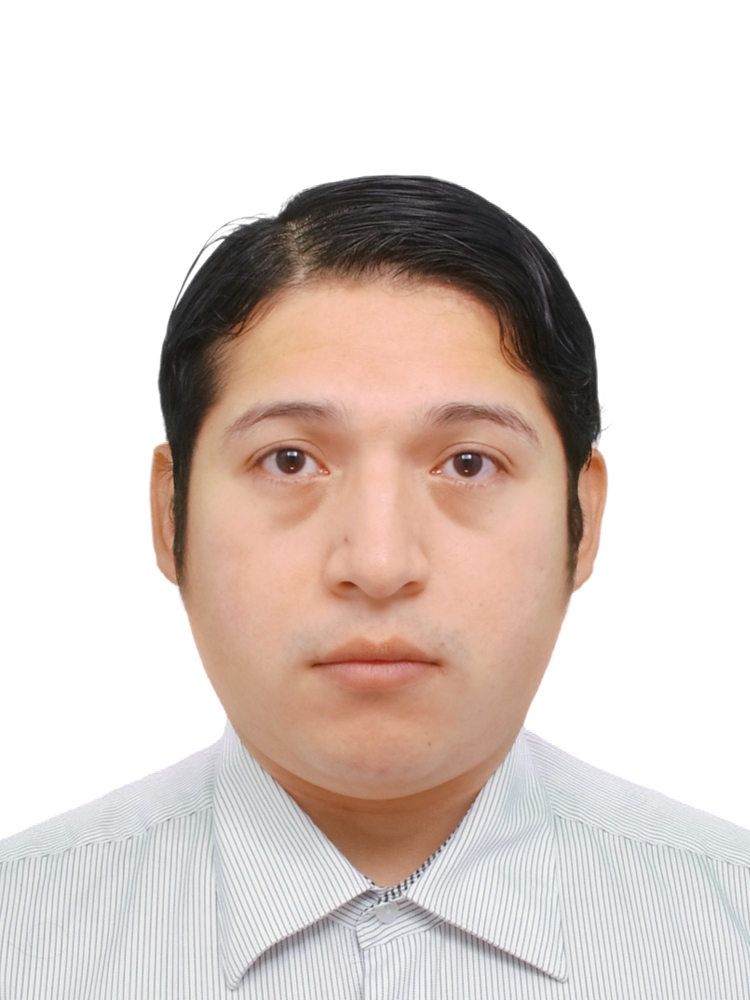}}]{Saúl Alexis Heredia Pérez} received the M.Sc. degree in Computer Science and Engineering from the National Autonomous University of Mexico in 2018 and the Ph.D. degree in Mechanical Engineering from The University of Tokyo in 2020. He is currently an Assistant Professor with the Center for Disease Biology and Integrative Medicine (CDBIM), Graduate School of Medicine, The University of Tokyo, Japan. From 2020 to 2023, he was an engineer with Sony Group Corporation. His research interests include virtual reality simulation, digital twins, and robot autonomy, with a particular focus on high-fidelity surgical simulation and autonomous robotic systems.
\end{IEEEbiography}%

\begin{IEEEbiography}[{\includegraphics[width=1in,height=1.25in,clip]{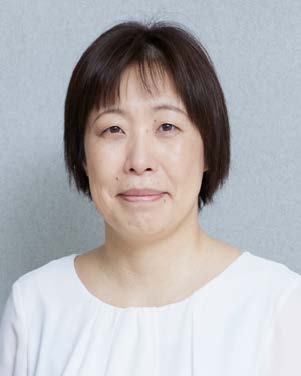}}]{Kanako Harada} (Member, IEEE) received the M.Sc. degree in engineering from The University of Tokyo in 2001 and the Ph.D. degree in engineering from Waseda University in 2007. She is currently a Professor with the Center for Disease Biology and Integrative Medicine (CDBIM), Graduate School of Medicine, The University of Tokyo, Japan. She also holds positions with the Department of Mechanical Engineering and the Department of Bioengineering, Graduate School of Engineering. Before joining The University of Tokyo, she held positions with Hitachi Ltd., the Japan Association for the Advancement of Medical Equipment, and Scuola Superiore Sant’ Anna in Italy. Her research interests include surgical robotic systems, robotic automation for biomedical applications, and regulatory science.
\end{IEEEbiography}%
\hfill

\end{document}